\documentclass{article} % For LaTeX2e
\usepackage{iclr2023_conference,times}

\usepackage[utf8]{inputenc} % allow utf-8 input
\usepackage[T1]{fontenc}    % use 8-bit T1 fonts
\usepackage{hyperref}       % hyperlinks
\usepackage{url}            % simple URL typesetting
\usepackage{booktabs}       % professional-quality tables
\usepackage{amsfonts}       % blackboard math symbols
\usepackage{nicefrac}       % compact symbols for 1/2, etc.
\usepackage{microtype}      % microtypography
\usepackage{xcolor}         % colors

\usepackage[pdftex]{graphicx}
\usepackage{caption}
\usepackage{subcaption}
\usepackage{hyperref}
\hypersetup{
    colorlinks=true,
    citecolor=black,
    linkcolor=black,
    urlcolor=blue,
    }
\usepackage{algorithm}
\usepackage{algpseudocode}
\usepackage{amsmath}
\usepackage{xspace}

\usepackage{textgreek}
\usepackage[normalem]{ulem}

\mathchardef\mhyphen="2D

\title{A Multiagent Framework\\
for the Asynchronous and Collaborative\\
Extension
of %Large-scale 
Multitask %Dynamic
ML Systems}

\author{%
  Andrea Gesmundo \\
  Google Research \\
  \texttt{agesmundo@google.com} \\
}

\begin{document}

\maketitle

\begin{abstract}
The traditional ML development methodology does not enable a large number of contributors, each with distinct objectives, to work collectively on the creation and extension of a shared intelligent system.
Enabling such a collaborative methodology can accelerate the rate of innovation, increase ML technologies accessibility and enable the emergence of novel capabilities.
We believe that this novel methodology for ML development can be demonstrated through a modularized representation of ML models and the definition of novel abstractions allowing to implement and execute diverse methods for the asynchronous use and extension of modular intelligent systems.
We present a multiagent framework 
%that enables the exploration of the proposed methodology 
for the collaborative and asynchronous extension of dynamic large-scale multitask systems.

\end{abstract}

\section{Introduction}

Traditional large-scale software systems enable teams of hundreds or thousands
of software engineers to work collectively on a single software artifact, through decomposition of the problem into many sub-problems, and through well-defined abstraction boundaries.
We currently lack the ability to have thousands of Machine Learning (ML) engineers and researchers collectively contribute to the continual extension of a single intelligent system.
The presented work aims to define a methodology for the automatic incorporation of new tasks and knowledge into a single running system, and the asynchronous and continual development of distinct methods each capable of extending the system in both capabilities and structure.
We see a direction where many individuals or teams can all contribute to the continuous improvement of a single intelligent system that is suited to a growing number of tasks and modalities, and that incorporates the learning and knowledge facilitated by the many people working on the same system each pursuing distinct objectives.
Each user of the system can benefit from the knowledge, methods and structures already embedded in the system, and in turn they can contribute to extend the system while pursuing their objectives.
In addition to enabling thousands of engineers and researchers to contribute to a single system, 
it may even be possible to have tens of millions of people without ML knowledge to contribute enriching a single ML system, by providing new tasks and data while applying the skills that have been taught to the system by others.

We present a multiagent framework that enables the exploration of the proposed methodology for collaborative and asynchronous extension of large-scale multitask dynamic ML systems.

\begin{figure}
  \centering
  \includegraphics[width=0.97\linewidth]{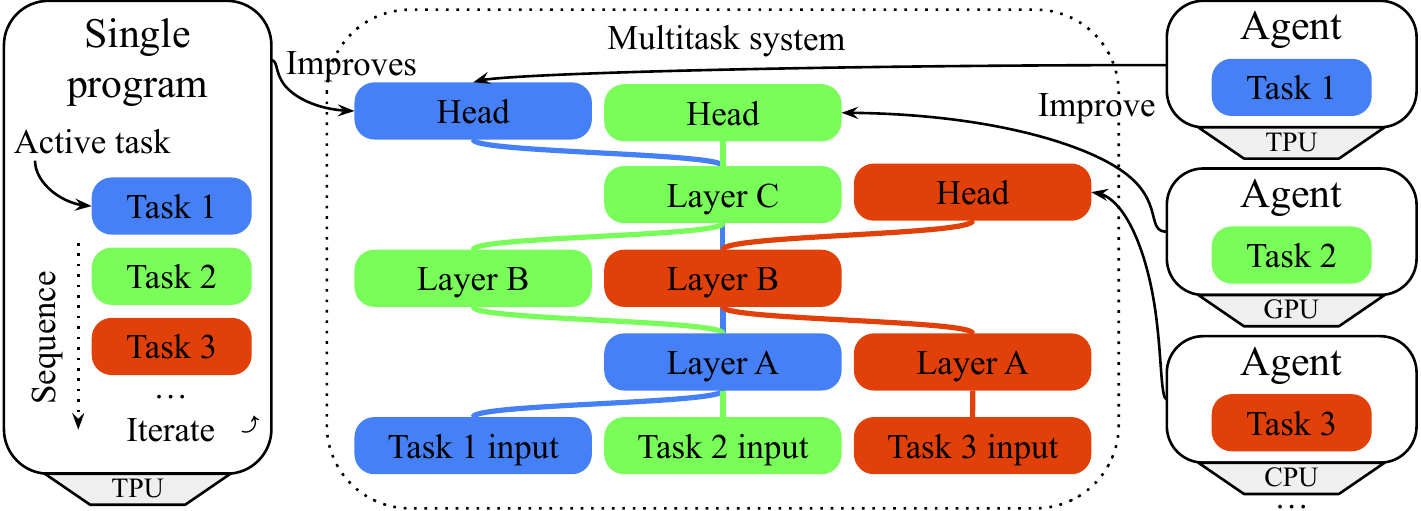}
  \caption{
  Graphical representation of the single program sequential execution (left) and the proposed multiagent parallel execution (right). The single program iterates over the set of tasks and improves one model/path at the time. Instead, multiple agents can be run in parallel, each works asynchronously to improve a substructure within the multitask system. In the example represented, each agent searches for an improved model for the assigned task. 
  }
\label{fig:mut}
\end{figure}

\section{Method}
\label{section:method}
We introduce the proposed framework 
% in the context
as an extension
of the continual ML development methodology
and the $\mu$2Net+ method introduced by \cite{Gesmundo2022munet3}.
$\mu$2Net+ is a method capable of evolving
% a Visual Transformer model \citep{Dosovitskiy2021AnII}
any modularizable ML model, such as deep neural networks, into a multitask ML system solving an unbounded set/stream of tasks.
This method is based on an evolutionary approach that iteratively samples and mutates models by applying different types of mutation actions: architectural changes (e.g. insert or remove layers), knowledge transfer selection (e.g. choose from which other model transfer parameters and which subset to fine-tune), and hyperparameter tuning (e.g. select piece-wise constant schedules for the hyperparameters of each generated model and of the evolutionary method).
Every mutated model is trained and scored on a single task, and can be subject to further mutations in future iterations.
The best performing model for each task is retained in the multitask ML system as part of a connected network of shared layers.
% , as the one represented in Figure~\ref{fig:graph}.
%and whose evolutionary process is represented in Video \href{https://youtu.be/Ld9gfmJT6Ig}{https://youtu.be/Ld9gfmJT6Ig}.
For details of the $\mu$2Net+ method and its sequential execution refer to \cite{Gesmundo2022munet2} and \cite{Gesmundo2022munet3}.

In prior work, the $\mu$2Net+ method has been applied with a \textbf{single program} sequential execution.
This constrained the method to search for an improved model/path for only one task at the time, the \emph{active task}.
% The single program execution searches for a better scoring model for each task by activating each task one at the time. 
New candidate models for the active task are generated, trained and scored while all other models composing the multitask system are immutable.
% all the model for the inactive tasks immutable.
With \emph{task-set iteration} we refer to the process of performing one active task iteration on each available task.
Multiple task-set iterations can be performed to extend the evolutionary/training process and provide more opportunities for cross-tasks knowledge transfer.

The framework proposed in this paper allows for multiple programs, \textbf{agents}, to jointly extend a shared multitask ML system.
Each agent can asynchronously create/extend substructures within the system.
In the instantiation of the multiagent framework presented by this publication, each agent is assigned a single task and applies the $\mu$2Net+ method with the objective of finding an improved model for the assigned task.
In this context, a task-set iteration can be achieved by having each agent perform one task iteration on the assigned task in parallel.
Notice that, even with the asynchronous execution, the properties of the $\mu$2Net+ method are retained. For example, it sill guarantees immunity against \emph{catastrophic forgetting} (the immutability of parent models is still guaranteed during the mutation and training of child models), \emph{negative transfer} (the unconstrained choice among a wide range of prior knowledge is sill available) and \emph{gradient interference} (the parameters of components unfrozen for training/fine-tuning still receive gradients for only one source within each generation).
The following paragraphs detail additional properties introduced by the proposed multiagent framework.

\paragraph{System scaling}
The proposed framework enables the possibility to achieve system scaling through agent parallelism.
Different scaling techniques are commonly applied to increase the compute resources accessible to ML models with the aim of increasing their size or accelerate their training on bigger datasets.
\emph{Data parallelism} techniques aim to replicate the model on multiple machines while synchronizing gradients across replicas \citep{Bottou2010LargeScaleML,Li2014ScalingDM}.
\emph{Model parallelism} techniques allow to parallelize the execution of large layers by partitioning their parameters across parallel hardware (i.e. multiple accelerators) requiring to distribute/gather the input/output tensors of each layer shard \citep{Dean2012LargeSD,Krizhevsky2014OneWT}.
\emph{Pipelining} techniques allow to place continuous sequences of layers on different machines but require extra communication between the pipeline stages and idle machine cycles for stages waiting for their input being propagated in both forward and backward training phases \citep{Huang2019GPipeET,Narayanan2019PipeDreamGP}. 

The proposed multiagent framework allows for an additional dimension of scaling that we can refer to as \emph{agent parallelism}.
In practice, more compute resources can be employed to accelerate the training throughput by running in parallel multiple agents.
Agent parallelism can be applied jointly with any of the other scaling technique.
Notice that, agent parallelism does not require any communication overhead since each agent works asynchronously. Each agent only needs to update its working image of the overall system state on a need basis. 
In the implementation proposed with this work, each agent updates their system state image by reading it from disk at the beginning of each evolutionary iteration as structures generated by other agents could be sampled as parent models.
% In multiagents experiments executed on TPUv4 with Megacore configuration optimized for dense computation (i.e. repurposing resources normally allocated for sparse computation),
% % reported in Section~\ref{section:experiments}, 
% we have recorded an average of
% 98.3\%
% % 99.2\%
% % 97.06\% TODO take more samples
% active TPU duty cycles, that normally would not be expected to be achievable with the common scaling techniques due to the additional synchronization required.
The properties of system scaling through task parallelism are further analyzed in Sections~\ref{section:experiments} and~\ref{section:experiments2}.
Notice that, the 4 different scaling techniques mentioned in this section are not exclusive, rather any subset of those techniques can be combined \citep{Park2020HetPipeEL}.

\paragraph{Method heterogeneity}
In the described instantiation of the proposed framework, all agents apply the same $\mu$2Net+ method to tasks that all have the same image classification framing.
Though, in general, the proposed multiagent framework supports agents differing in any aspect of the method such as: reward function, input/output task modalities, root model, space of generatable architectures, mutation actions and hyperparameters search space or even search method, for example employing Reinforcement Learning or Bayesian methods instead of an Evolutionary approach.

\paragraph{Hardware heterogeneity}
The general principles of the proposed framework and the particular implementation we provide\footnote{Source-code available at \href{https://github.com/google-research/google-research/tree/master/muNet}{https://github.com/google-research/google-research/tree/master/muNet}}, allow for each agent to be run as a distinct program on a different type of hardware.
In a preliminary experiment, we have tested the hardware heterogeneity capability by generating a multitask system solving three tasks with three distinct agents each run on different hardware type (Figure~\ref{fig:mut}).
The first agent was run on a TPUv3 machine with 8 cores (thus sampling and training 8 models in parallel).
The second agent was run on a Tesla P100 GPU machine with 1 core.
While, the third agent was run on a single CPU.
Furthermore, each agent was executed on a different data-center.
This capability is achieved with no software customization, since each agent runs as a distinct program and synchronizes only via the sharded system state stored on disk, i.e. cloud storage.
This hardware heterogeneity capability allows to modulate the resources assigned to each agent according to availability and requirements of different users. The hardware allocation and requirements can also change through time and their cost can be dynamically factored into the reward function of each agent as described in \citet{Gesmundo2022munet1,Gesmundo2022munet3}.

\paragraph{Collaborative ML development}
The proposed framework provides the abstractions enabling the novel collaborative ML development methodology.
This framework allows multiple users, each aiming to solve a different research or applied problem, to introduce an agent designed to solve the task at hand with improved quality and efficiency by leveraging the knowledge and methods already embedded in the shared multitask dynamic ML system.
In turn, each new agent enriches the ML systems with new knowledge, structures and capabilities available to future users.

\begin{figure}[t]
  \centering
%   \fbox{\rule[-.5cm]{0cm}{4cm} \rule[-.5cm]{4cm}{0cm}}
\text{Multitask Character Classification Benchmark}\par%\medskip
  \includegraphics[width=0.5\linewidth]{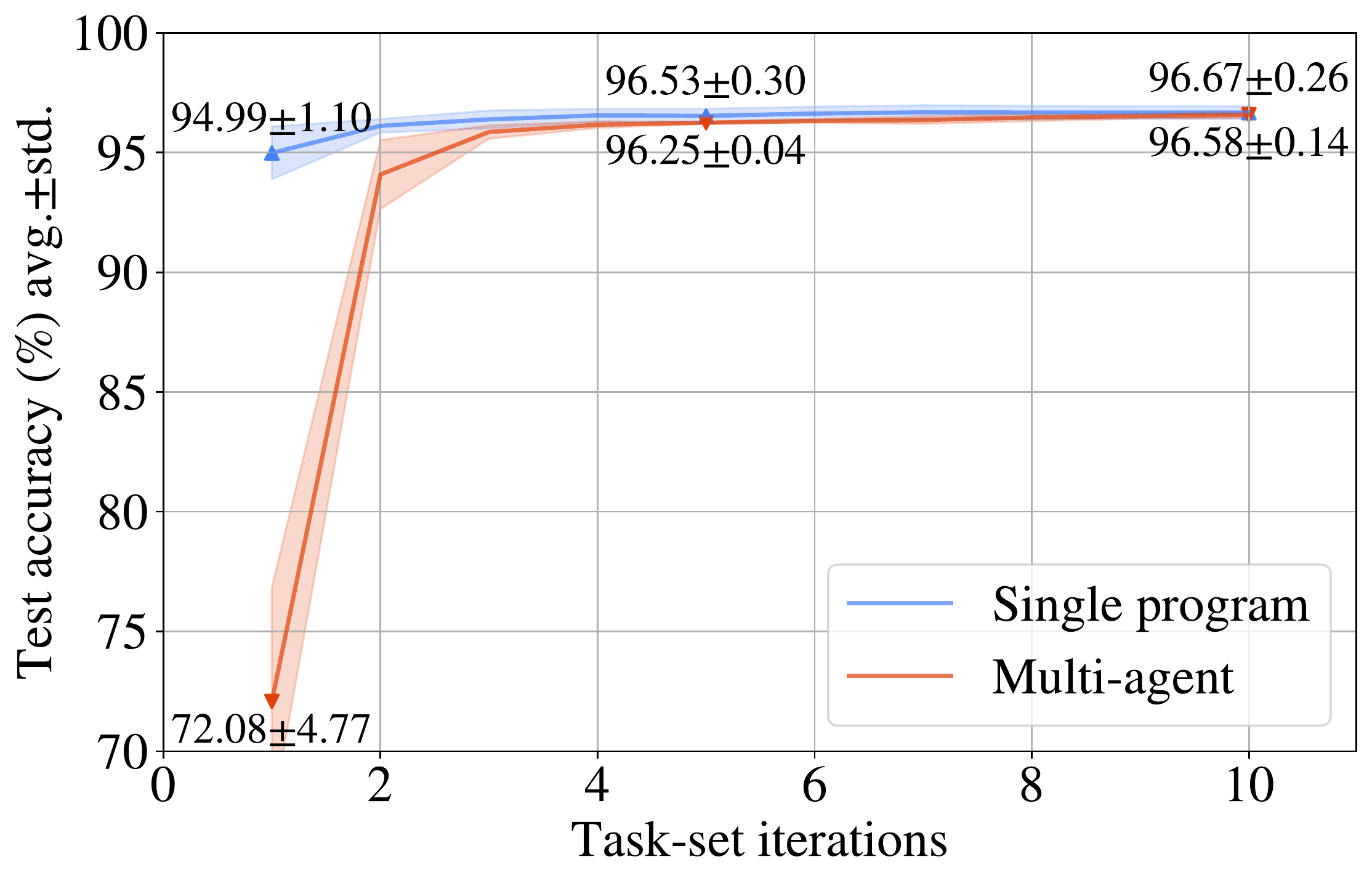}%
  \includegraphics[width=0.5\linewidth]{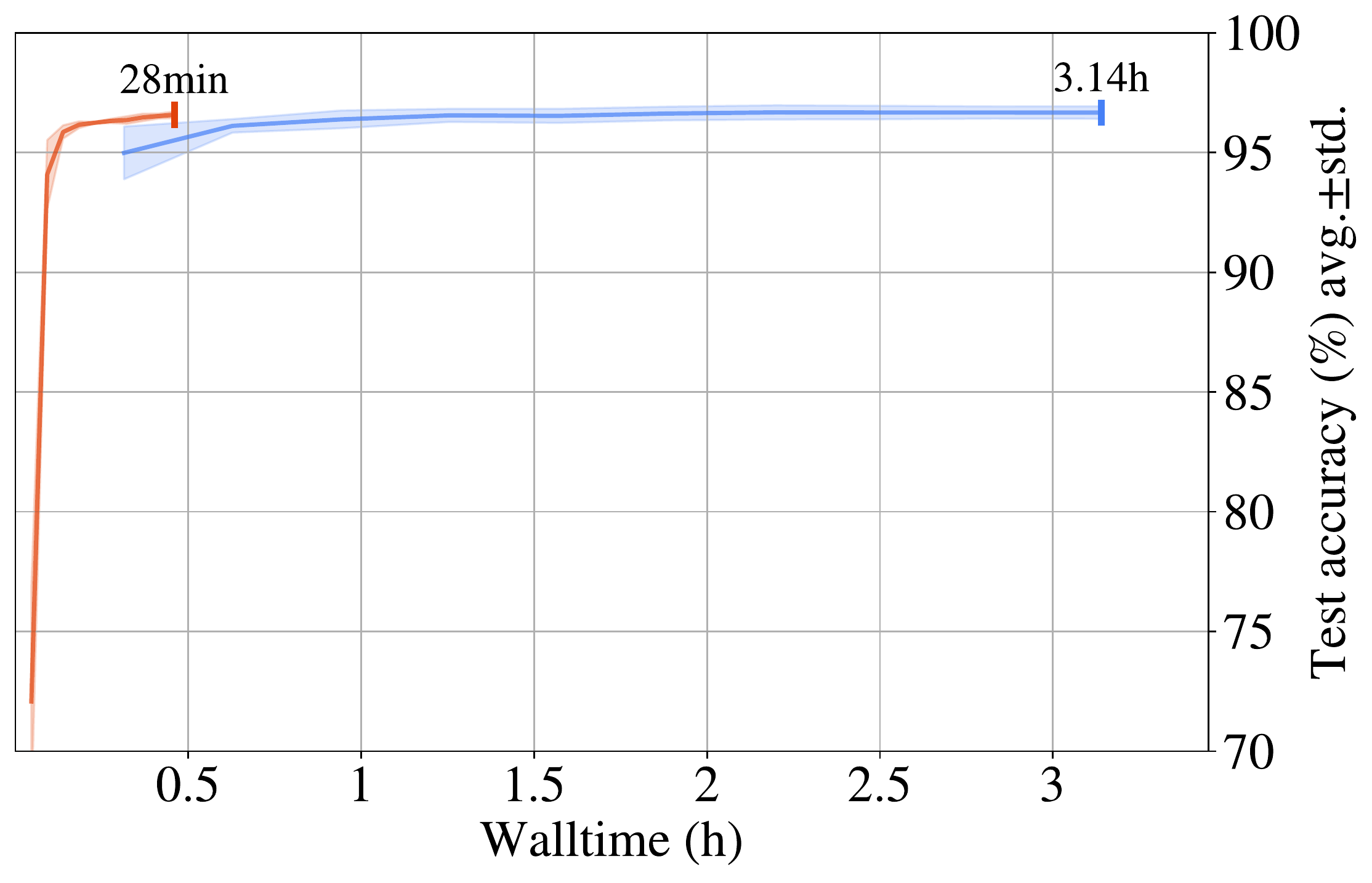}
\text{Visual Domain Decathlon Benchmark}\par%\medskip
  \includegraphics[width=0.5\linewidth]{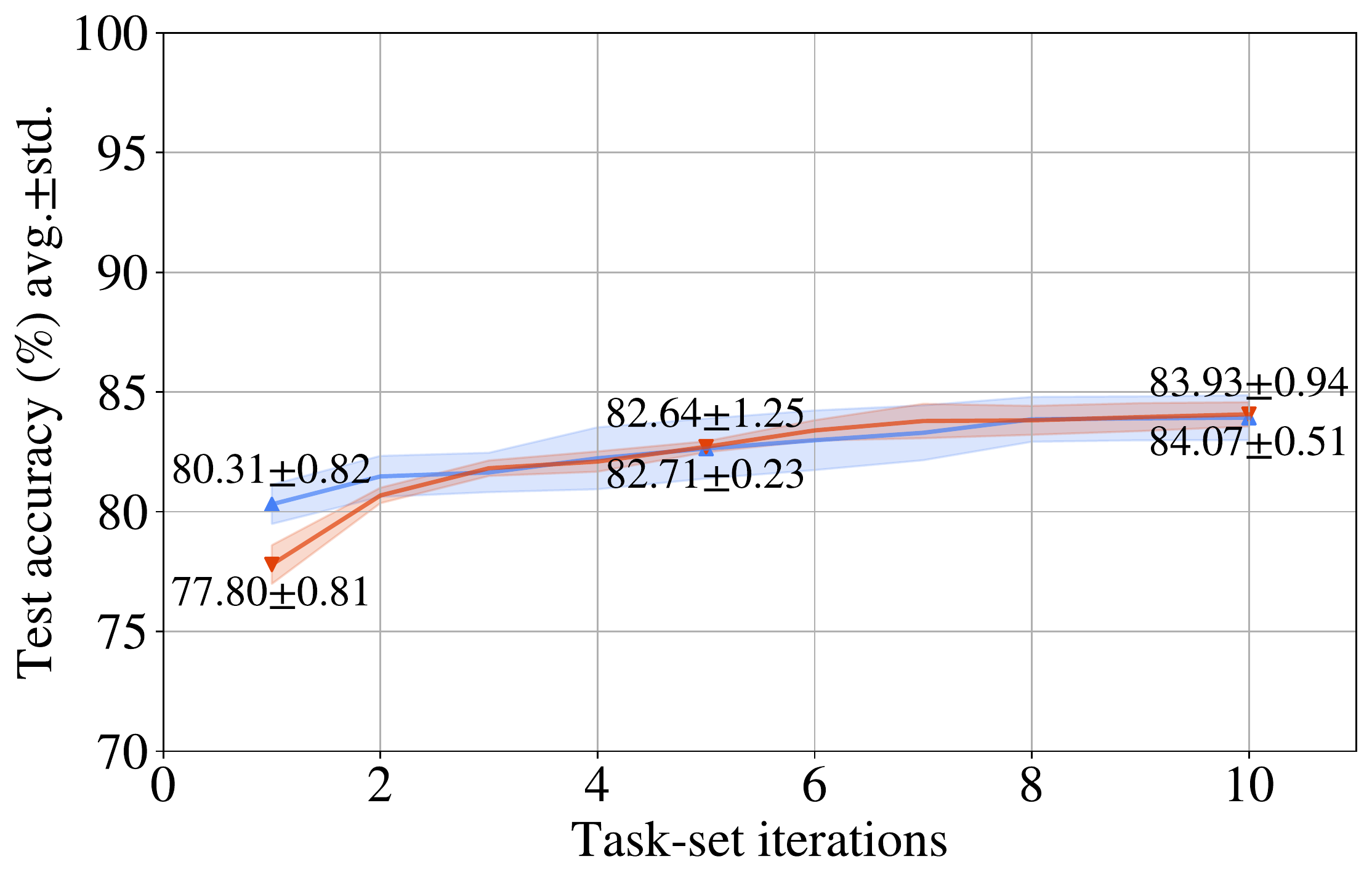}%
  \includegraphics[width=0.5\linewidth]{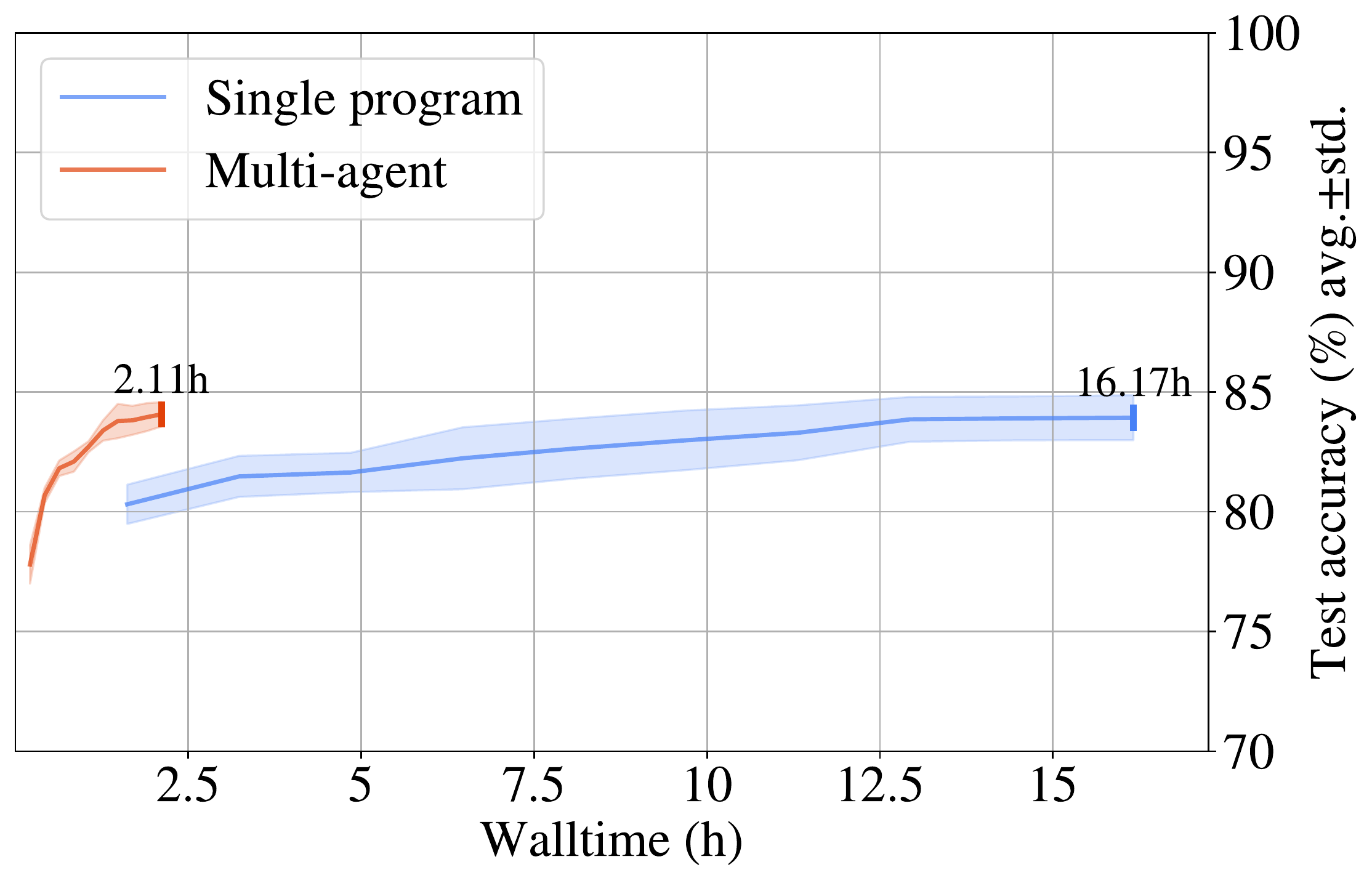}
  \caption{Comparison between the sequential (single program) and parallel (multiagent) executions of the $\mu$2Net+ method generating multi-task systems solving jointly the tasks of the Multitask Character Classification Benchmark and the Visual Domain Decathlon Benchmark.
  The horizontal axis of the left plots measures he number the task-set iterations, while the horizontal axis of he right plots measures the time since the start of he experiment.
  The vertical axis of all plots measures the test accuracy averaged across all the tasks composing each benchmark.
  The curves display the aggregated results of the experiment repetitions.
  Vertical coordinate of the curves represent the test accuracy averaged across the corresponding experiment repetitions.
  The shaded areas enveloping the curves represent the standard deviation of the test accuracy compute across repetitions.
  The numerical values of the test accuracy average and standard deviation achieved after 1, 5 and 10 task-set iterations are reported in the left plots.
  The time to complete 10 task-set iterations is displayed in the right plots.
  }
\label{fig:comp-exp}
\end{figure}

% \section{Comparison Experiments}
\section{Experiments}
\label{section:experiments}

This section reports an empirical analysis of the properties of the proposed multiagent framework employed as a scaling technique.
We compare against the baseline ``single program'' sequential execution.

To validate the generality of the analyzed properties,
the experiments are repeated on two benchmarks: the Multitask Character Classification Benchmark (MCCB) and the Visual Domain Decathlon Benchmark (VDDB).
The MCCB is defined by a set of eight character classification tasks similar in domain type,
while VDDB is composed by ten tasks with larger datasets that have been explicitly selected to each represent as distinct domains as possible. Refer to Table~\ref{table:datasets} for tasks details and references.

The experiments representing the single program sequential execution apply the $\mu$2Net+ method with configuration equivalent to that described in \cite{Gesmundo2022munet3}.
In summary, each task iteration is composed by 4 generations,
during each generation 8 models are sampled and trained in parallel,
each sampled model is trained for 4 epochs on the current active task where each epoch is capped to a max of 51200 training samples (equivalent to 100 training batches as each batch is composed by 512 samples).
The root model architecture for MCCB is a ViT Tiny model capped to 3 transformers layers and for VDDB is a full ViT Base model with 12 transformer layers.
The MCCB experiments are repeated 5 times to measure metrics variance, while the VDDB experiments are repeated 3 times due to the higher experiment cost.
The root models parameters are loaded from a checkpoint pretrained on ImageNet-21k as described in \citet{Steiner2021HowTT}.
The only difference, compared with the configuration used by \citet{Gesmundo2022munet3}, is that the cost scale factor is fixed to 1 and the number of task-set iterations is set to 10 for both benchmarks.

The experiments representing the multiagent execution use the same $\mu$2Net+ configuration and equivalent training compute budget in terms of total training steps per tasks.
The only difference, compared with the single program execution, is that each task-set iteration is executed by multiple agents in parallel.
In more details, we create one agent per task,
each agent performs one task iteration on the assigned task by applying the same $\mu$2Net+ evolutionary method to find a better scoring model for its task.
Each agent is executed asynchronously and in parallel to the other agents.
The only constraints imposed is that agents cannot start the $n^{th}$ task iteration until all the agents have completed the $(n-1)^{th}$ iteration.
Notice that, agents iteration time may vary since the $\mu$2Net+ method allocates a bigger train budget to tasks with bigger dataset and thus a longer train epoch.
Therefore, in this specific application, the end-to-end completion time is bottlenecked by the speed of the slowest agents.
To avoid inefficiencies, the faster agents release the training resources as they wait for the start of the next iteration.

The results of these comparison experiments are displayed in Figure~\ref{fig:comp-exp}.
In both benchmarks it is noticeable that the multiagent execution achieves a significantly lower accuracy after the first task-set iteration.
% This is justifiable if we consider that 
During the very first iteration, the parallel agents can generate models that transfer knowledge only from the root model.
This leads to reduced quality compared to the sequential approach that allows for knowledge to be transferred also from all the tasks that are preceding in the sequence. % or the assigned task.
In few iterations, the gap between the quality achieved by the two methods is reduced until the quality deltas are withing statistical noise range.
The initial gap on the MCCB is significantly larger, this could be justified considering that the MCCB tasks are similar in domain (all character classification tasks) so they can be expected to suffer more from reduced knowledge transfer in the initial iteration compared to VDDM tasks that have explicitly selected to represent a set of domains as different as possible.

The proposed multiagent execution achieves a sharp reduction in execution time (Figure~\ref{fig:comp-exp} right).
The MCCB parallelized experiments are 6.64 times faster than the sequential counterparts.
Notice that, MCCB is composed by 8 tasks, and its experiments are parallelized with 8 agents. Therefore, the upper bound for speed improvement is 8 times faster. This can be achieved by constraining all the agents to use the same training budget, rather than assigning a training budged proportional to each task train set.
The VDDB experiments result 7.66 times faster, and its speed-up upper bound is 10.

\section{$\mu$3Net system extension}
\label{section:experiments2}
\citet{Gesmundo2022munet3} introduces a \emph{continual development methodology} that allows to continuously extend dynamic multitask systems.
This methodology has been employed to extend the Mutant Multitask Network ($\mu$2Net) \citep{Gesmundo2022munet2} with additional tasks/knowledge and method improvements, resulting into an improved network referred to as $\mu$2Net+.
In this section, we further demonstrate the continual development methodology by extending once more the $\mu$2Net+ system into a Mutant Multitask Multiagent Network ($\mu$3Net).
This is achieved by continuing the training/evolution of the $\mu$2Net+ system generated by the experiments reported in \citet{Gesmundo2022munet3} by applying the parallel multiagent execution instead of the sequential one.
All the metaparameters of the method such as hyperparamters search space, reward function or compute budget per task iteration are left unchanged (for details refer to \citet{Gesmundo2022munet3}). Since the system solves jointly 124 tasks, 124 agents are executed in parallel.
Each agent evolves and fine-tunes models/paths to maximize the reward achieved on one assigned task.
Each agent performs 16 task-iteration. During each task-iteration, 4 generations of new models/paths are sampled, trained and scored. During each generation, 4 paths are sampled and trained in parallel.
Each agent is executed one TPUv4 machine in MegaCore configuration, resulting in 4 TPU chips assigned to each agent.

Within the context of the $\mu$Net line of research, this experiment demonstrates the further extension of the system and method by applying the continual development methodology. While, considered in isolation, this experiment mainly demonstrates the ``system scaling'' property of the proposed multiagent framework.
Following the measurement method defined in Section~\ref{section:experiments}, the experiment speed-up is computed as the ratio between the time required to complete a task-set iteration sequentially on all the 124 tasks and the time it takes for the slowest agent to complete a task iteration.
The average speed-up factor achieved by the multiagent parallel execution results approximately equal to 26.1 times.
This can be regarded as a significant speed-up considered that a 4 weeks experiment can be reduced to approximately one day.
Although this is far from the maximum 124 times speed-up that could be achieved by modifying the method to achieve equivalent compute budget across agents.
Instead, applying the logic defined in \citet{Gesmundo2022munet3}, agents assigned to tasks with small datasets will train each new model/path on few hundred examples (e.g. VTAB-1k tasks), while agents assigned to tasks with larger datasets will train on hundred of thousands of examples (e.g. ImageNet).

\begin{figure}[t]
  \centering
\text{Segment 5: $\mu$3Net}\par%\medskip
  \includegraphics[width=1.\linewidth]{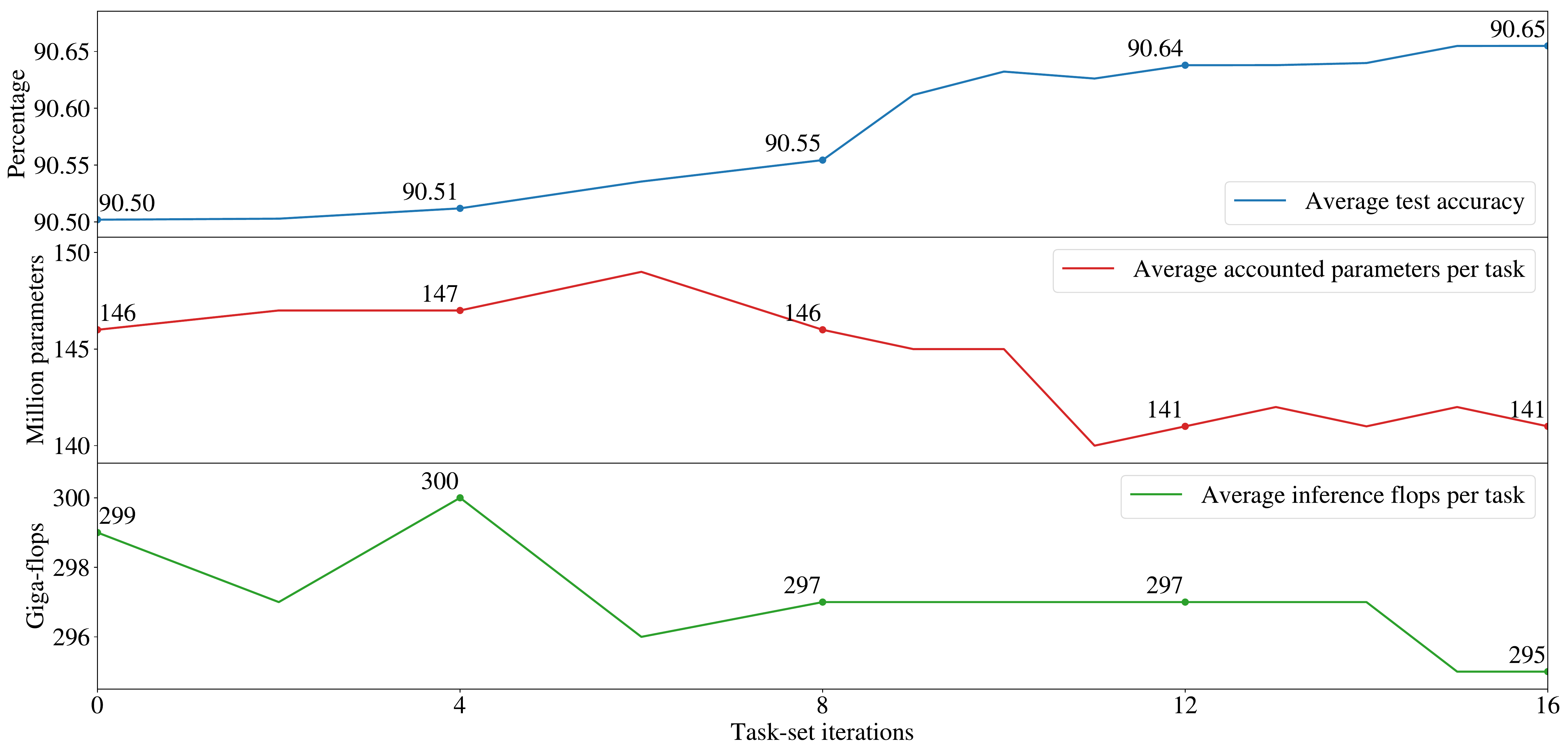}
  \caption{
Displays measures of the 3 reference metrics through the sequence additional 16 task-set iterations:
  1)~Average test accuracy achieved on the 124 tasks, the test accuracy is computed on a test set that does not overlap with the training set and validation set (whose validation accuracy is used for the reward computation) to avoid overfitting and selection bias.
  2)~Accounted parameters averaged across all the 124 models/paths composing the multitask system. This measure corresponds to the size cost factor included in reward function \citep{Gesmundo2022munet3}.
  3)~Compute cost measured as flops required to produce inference for one input sample, also averaged across all the models/paths composing the multitask system. This measure corresponds to the compute cost factor included in reward function \citep{Gesmundo2022munet3}.
We can refer to this as the 5th experiment segment, as it further extends the 4 segments described in \citet{Gesmundo2022munet3}.
  }
\label{fig:comp-exp}
\end{figure}

Figure~\ref{fig:comp-exp} reports the measures of the 3 main metrics considered by the reward function. Notice how the continuation of training/evolution for the additional 16 task-set iterations results into a small improvement on all 3 metrics.
We can observe an improvement in both validation accuracy (used by the reward function) and test accuracy (held out) their averages across the 124 tasks change respectively from 92.24\% to 92.36\% (validation) and from 90.50\% to 90.65\% (test).
These quality gains appear to be the result of further cross-task knowledge transfer.
The quality gains are prevalent on the most recently introduced tasks \citep{Gesmundo2022munet3}.
The reduction in ``average accounted parameters per tasks'' is mostly due to a reduction in average number of ``transformer layer'' components used per model/path: from 23.62 to 23.56.
While the reduction in ``average inference flops per task'' is mostly due to an increase in number of models/paths using the lower input resolution option: from 38 models to 40.

This experiment segment includes an additional method extension that we mention for completeness, although it does not appear to have had a significant impact on the results.
This method extension consists in the addition of a mutation action that allows to add new randomly-initialized trainable ``transformer layer'' components above the 24th transformer layer.
All the models/paths currently in the systems have been evolved from a single ViT-Large root model, as described in \citet{Gesmundo2022munet2}.
The root model has 24 transformer layers.
The set of mutations available so far allowed to incrementally remove transformer layers to offer an additional way to reduce the size and compute cost factors of the scoring function.
Thus, it was not possible so far to generate models bigger than the root model.
The ``layer addition'' action has been introduced to enable the generation of larger scale models that appear to be required to achieve state-of-the-art performance on complex tasks with large datasets such as ImageNet.
During the \href{https://www.youtube.com/watch?v=eUfIa-HqgeI}{evolutionary process}, we observe that 7 new ``best scoring'' models/paths with more than 24 layers have been added to the system.
But, by the end of the 16th task-set iteration only one of such models remains.
Thus we conclude that the introduction of the ``layer addition'' action has no significant effect so far and the evolution of significantly larger models may require more evolutionary iterations and possibly the introduction of more complex tasks such as ImageNet-21k or JFT-300M on which the ViT root models were pre-trained \citep{Steiner2021HowTT}.

\begin{figure}[h]
\centering
\includegraphics[width=1.\linewidth]{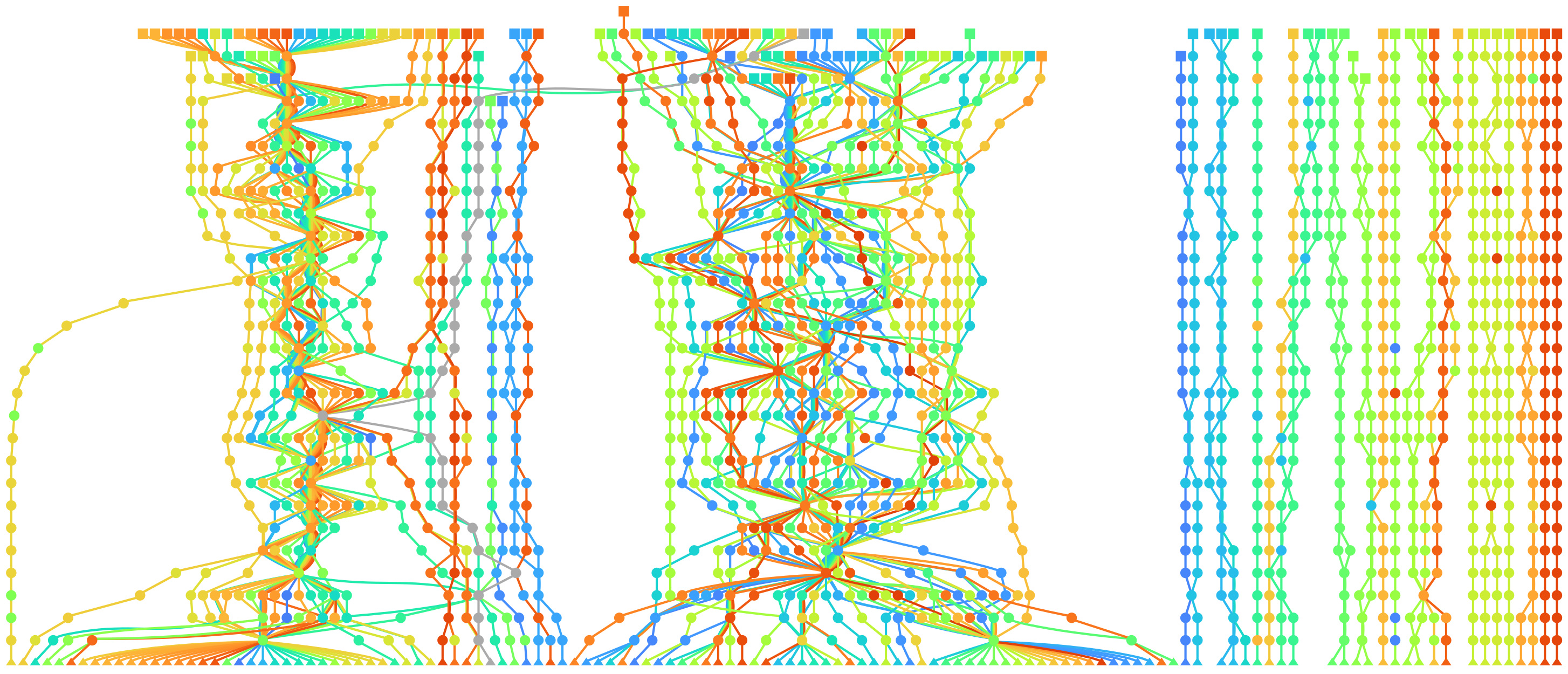}
\caption{Graph representing the architecture of the $\mu$3Net multitask system solving jointly 124 image classification tasks generated by the experiment described in Section~\ref{section:experiments2}.
Each task is identified with a unique color.
Bottom triangular nodes represent the data input of each task.
Top rectangular nodes represent the head layer of each task.
Each sequence of edges of the same color connecting a task input to its head defines the layers sequence composing the model for each task.
Each path traverses a sequence of round nodes representing ViT Large internal layers in the following order from bottom to top: patch embedding, class token, position embedding and a variable number of transformer layers.
Internal nodes are represented with the color of the task on which the parameters of the corresponding layer were trained/fine-tuned last.
The gray nodes have not received gradient updates from any of the 124 tasks and still carry the parameters of the root model that was loaded from a checkpoint of the ViT Large model pretrained on the ImageNet-21k dataset as described in \cite{Gesmundo2022munet2}. Video [\href{https://youtu.be/eUfIa-HqgeI}{youtu.be/eUfIa-HqgeI}] displays an animation of the evolutionary process.
Notice that there is only 1 model/path retaining a 25th transformer layer at the end of the 16 task-set iteration. That is the model/path solving the \href{https://www.tensorflow.org/datasets/catalog/sun397}{sun397} task.
% \\
% The structure displays the formation of two main clusters of models/paths having a high degree of layers sharing.
% These are models selected for tasks that can achieve peak quality with a lower degree of knowledge specialization.
% Few models form smaller and disconnected clusters of paths.
% Notice that knowledge sharing is effective even for models/paths that result in a disconnected sub-graph, since ancestors models may have been trained on multiple tasks whose current best model has fully branched in a separate disconnected structure during the evolutionary process.
% Among these small clusters it is visible a high degree of sharing among related tasks.
% For example, the leftmost disconnected subgraph of 3 paths  contains only characters classification tasks: \emph{bangla}, \emph{devanagari} and \emph{binary\_alpha\_digits} (see Table~\ref{table:datasets} for datasets details).
% The majority of the disconnected subgraphs of 2 paths contain each a VTAB-Full task paired with the matching VTAB-1k "few-shots learning" version of the same task, from left to right: 1) \emph{clevr/count\_all} and \emph{clevr/count\_all}$_{1k}$,
% 2) \emph{dmlab} and \emph{dmlab}$_{1k}$,
% 3) \emph{dsprites/label\_orientation} and \emph{dsprites/label\_orientation}$_{1k}$,
% 4) \emph{eurosat} and \emph{eurosat}$_{1k}$,
% 5) \emph{resisc45} and \emph{resisc45}$_{1k}$,
% 6) \emph{smallnorb/label\_azimuth} and \emph{smallnorb/label\_azimuth}$_{1k}$.
}
\label{fig:graph-big}
\end{figure}

Figure~\ref{fig:graph-big} displays the models/paths composing the $\mu$3Net system after the additional 16 task-set iterations.
Table~\ref{table:datasets} reports the validation and test accuracy achieved on each task.

\section{Related work}

This paper contributes to a line of research aiming to define and demonstrate a novel ML research methodology and methods that enable the creation of dynamic large-scale multi-modal multitask intelligent systems that can be indefinitely and collaboratively extended: \citep{Gesmundo2022munet1,Gesmundo2022munet2,Gesmundo2022munet3}.

The proposed method is designed to learn an unbounded number of tasks in a continual learning fashion.
In such contexts it aims to learn each task with higher quality and efficiency by automating and optimizing the knowledge transfer among any subset of tasks that can provide useful knowledge to one another.
The proposed model is designed to be immune from common multitask learning pitfalls: catastrophic forgetting, gradients interference, negative transfer. 
Cross-task \textbf{transfer-learning}
% has gained popularity in the form of transfer learning from a model pre-trained on a large amount of data and then fine-tuned on a small %amount of data for a 
% related downstream task.
has gained popularity, especially through transfer learning from a model
pre-trained on a large amount of data for one or a few general tasks,
and then fine-tuned on a small amount of data for a related downstream task.
This approach has been shown to be very effective in a wide variety of problems
across many modalities, including
% for different modalities such as 
%natural 
language \citep{Devlin2019BERTPO,Raffel2020ExploringTL} and vision \citep{Dosovitskiy2021AnII,He2016DeepRL}.
The success of transfer-learning applications hinges on adequate prior knowledge selection to avoid typical \textbf{negative transfer} pitfalls \citep{Rosenstein2005ToTO,Wang2019CharacterizingAA}.
Common solutions rely on data or model selection techniques,
often putting emphasis on the efficiency of the exploration
\citep{Zhang2020ASO,Mensink2021FactorsOI}, also method aiming to automate knowledge selection at a layer level have been proposed \citet{Sun2020AdaShareLW}.
Transfer learning capabilities are critical for \textbf{multitask models}.
ML models trained jointly on multiple tasks 
can be affected by \textbf{gradients interference} if any subset of parameters receive gradients jointly from multiple sources \citep{Chen2018GradNormGN,Yu2020GradientSF}, and by \textbf{catastrophic forgetting} of prior knowledge as new tasks are learned \citep{McCloskey1989CatastrophicII,French1999CatastrophicFI}.
These knowledge loss problems can be alleviated with weighted combination of tasks \citep{Liu2019LossBalancedTW,Sun2020ERNIE2A} and gradient transformation methods \citep{Chen2018GradNormGN,Sener2018MultiTaskLA,Kendall2018MultitaskLU}.
Stronger guarantees are provided by methods that compartmentalize task specific knowledge in dedicated parameter subsets \citep{Rebuffi2017LearningMV,Houlsby2019ParameterEfficientTL,Rusu2016ProgressiveNN,Rosenfeld2020IncrementalLT}.
Addressing catastrophic forgetting and identifying what subset of parameters/knowledge that is beneficial to share with each task is also critical for \textbf{continual learning} or life long learning methods
\citep{McCloskey1989CatastrophicII,French1999CatastrophicFI,Ramesh2022ModelZA}.

The proposed method relies on an evolutionary approach to jointly search the spaces of models architectures, hyperparameters, and prior knowledge selection while optimizing for an possibly multi-factor non-differetiable reward function.
The automation of \textbf{hyperparameter tuning} has been commonly addressed with Bayesian optimization \citep{Srinivas2010GaussianPO,Bergstra2011AlgorithmsFH,Snoek2012PracticalBO},
evolutionary methods have also been explored for this purpose
\citep{Jaderberg2017PopulationBT,Zhang2011EvolutionaryCM}.
Hyperparameters tuning can be considered related to the \textbf{neural architecture search} (NAS), as architectures can be defined by the selection of a sequence
of architectural hyperparameters.
Initially, NAS methods have been based on reinforcement learning techniques
\citep{Zoph2017NeuralAS} but also sample efficient evolutionary approaches have been proposed \citep{Real2019RegularizedEF,Maziarz2018EvolutionaryNeuralHA}.
Parameter-sharing based NAS methods aim to reduce the typically high training cost \citep{Pham2018EfficientNA,Liu2019DARTSDA,Kokiopoulou2019FastTA}.
Optimization for multi-factor quality/cost trade-offs have been explored \citep{Tan2019MnasNetPN}.

The proposed method is capable to dynamically extend the system, adding capacity or novel structures in an unconstrained fashion.
A few methods have been proposed to achieve \textbf{dynamic architecture extensions} \citep{Chen2016Net2NetAL,Cai2018EfficientAS}, some also focusing on an unbounded stream of tasks \citep{Yoon2018LifelongLW,Yao2020OnlineSM}, or achieving immunity from catastrophic forgetting  \citep{Rusu2016ProgressiveNN,Li2018LearningWF,Li2019LearnTG,Rosenfeld2020IncrementalLT}.

The proposed method is sparsely activated, thus the unbounded growth of knowledge and parameters is decoupled from the growth of computational cost.
The growth in capabilities of state of the art models often requires growth in terms of trainable parameters \citep{Kaplan2020ScalingLF}.
\textbf{Sparse activation} techniques at sub-layer level \citep{Shazeer2017OutrageouslyLN,Du2021GLaMES} or network route level \citep{Fernando2017PathNetEC} allow to decouple model size growth from compute cost.
This is achieved by integrating % in the model
a \textbf{routing technique} that selects the appropriate subset of parameters storing the most relevant knowledge for each task, sample or token/patch.

The ability of jointly solve a \textbf{large amount of tasks} is commonly associated with progress toward Artificial General Intelligence (AGI).
Advancements in scaling language models \citep{Brown2020LanguageMA,Thoppilan2022LaMDALM} allowed to achieve novel discourse, reasoning and zero/few shot learning capabilities that can be applied to new tasks without/minimal additional training.
Recent work aims to extend these achievements beyond text modality by defining static architectures for an extended subset of modalities \citep{Alayrac2022FlamingoAV,Reed2022AGA}.
These are a few examples of the ML models contributing to the line of research achieving incremental milestone toward AGI.
Though, each model is trained from scratch with considerable resources consumption.
The introduction of abstractions allowing to modularize, dynamically extend and reuse these large models may contribute to accelerate the rate of innovation. % and increase accessibly.
% TODO extend with Berner from P3

% TODO mulltimodal extensino
% Frozen, flamingo
% Pix2seq example of task framing extensions

% TODO Ability to optimize for hybrid functions 
% Sun2020AdaShareLW  MNAS ...

%  TODO Other missing citations include: (b) selection and mutation operators from evolutionary computation literature and c) prior work in dataset privacy.

% TODO multiagent frameworks.

% TODO 
%New type of parallelism other than data/model parallelism or pipelining. See above in paper 4

\section{Conclusion}
\label{section:conclusions}

This publication introduces a multiagent framework for the collaborative and asynchronous extension of large-scale multitask dynamic ML systems.
The reported empirical study demonstrates how the framework can be employed to achieve hardware heterogeneity and efficient scaling for a multitask system.
Future work can continue to build toward defining and studying the properties of extended ecosystems of agents.
In which, agents can implement distinct methods/objectives and can transfer knowledge across tasks with different input/output modalities.
Such novelties can be demonstrated with a continual development methodology, possibly
further extending large-scale dynamic multitask systems such as $\mu$3Net.
Source code and checkpoint of the $\mu$3Net system are publicly available at \href{https://github.com/google-research/google-research/tree/master/muNet}{https://github.com/google-research/google-research/tree/master/muNet}.

\vspace{10pt}

\clearpage
\begin{table*}[h]
\small
\caption{Datasets details (part 1 of 3).
For each dataset used in the experiments, this table reports:
1) accuracy achieved on the test and validations sets by the large scale multitask model generated by the experiments described in Section~\ref{section:experiments2},
2) dataset name indicative of the Tensorflow Datasets Catalogs identification string, linking to the corresponding catalog page,
3) train, validation and test data splits, represented with the \href{https://www.tensorflow.org/datasets/splits}{standard Tensorflow Datasets format} (``validation'' has been abbreviated as ``val''),
4) corresponding scientific publication reference.
Datasets are listed in the order of introduction into the %multitask
system.
\\\hspace{\textwidth}
Notes:
\\\hspace{\textwidth}
[1] The test split of the \href{https://www.tensorflow.org/datasets/catalog/imagenet_v2}{imagenet\_v2} dataset is used as validation set for \href{https://www.tensorflow.org/datasets/catalog/imagenet2012}{imagenet2012}. % following \citet{Steiner2021HowTT}.
\\\hspace{\textwidth}
[2] The test split of the \href{https://www.tensorflow.org/datasets/catalog/cifar10_1}{cifar10\_1} dataset is used as validation set for \href{https://www.tensorflow.org/datasets/catalog/cifar10}{cifar10}.
\\\hspace{\textwidth}
[3] The VTAB-full benchmark also includes the cifar100 task. Cifar100 has been introduced to the system as part of the initial benchmark.
\\\hspace{\textwidth}
[4] The definition for the VTAB standard and additional tasks has been sourced from \href{https://github.com/google-research/task_adaptation/tree/master/task_adaptation/data}{https://github.com/google-research/task\_adaptation/tree/master/task\_adaptation/data}.
\\\hspace{\textwidth}
[5] VTAB additional task, not included in the standard scoring set. These tasks were added to further scale the system and analyze transfer across related tasks.
}
\label{table:datasets}
\centering
\setlength\tabcolsep{1pt}
\hspace*{-13.32pt}
\begin{tabular}{lcccccc}
    \toprule
    &\multicolumn{2}{c}{Accuracy (\%)}
    & \multicolumn{3}{c}{Splits}                   \\
    \cmidrule(r){2-3}
    \cmidrule(r){4-6}
    Name & Val. & Test   & Train & Val. & Test  & Reference\\
    \midrule
    \midrule

\multicolumn{7}{c}{\textbf{Task-set A}} \\
\midrule

\href{https://www.tensorflow.org/datasets/catalog/imagenet2012}{imagenet2012}
&  78.54  &  86.66
& train & {\tiny \href{https://www.tensorflow.org/datasets/catalog/imagenet_v2}{imagenet\_v2}:}test$^{[1]}$ & val
& \citep{Russakovsky2015ImageNetLS}
\\
\href{https://www.tensorflow.org/datasets/catalog/cifar100}{cifar100}
&  96.83  &  94.70
& train[{\tiny:98\%}] & train[{\tiny98\%:}] & test
& \citep{Krizhevsky2009LearningML}
\\
\href{https://www.tensorflow.org/datasets/catalog/cifar10}{cifar10}
&  98.81  &  99.48
& train & {\tiny \href{https://www.tensorflow.org/datasets/catalog/cifar10_1}{cifar10\_1}:}test$^{[2]}$ & test
& \citep{Krizhevsky2009LearningML}
\\

\midrule
\multicolumn{7}{c}{VTAB-full benchmark$^{[3][4]}$} \\

\href{https://www.tensorflow.org/datasets/catalog/caltech101}{caltech101}
   &  98.67  &  96.17
& train[{\tiny:2754}] & train[{\tiny2754:}] & test
& \citep{FeiFei2004LearningGV}
\\
\href{https://www.tensorflow.org/datasets/catalog/dtd}{dtd}
&  82.63  &  82.18
& train & val & test
& \citep{Cimpoi2014DescribingTI}
\\
\href{https://www.tensorflow.org/datasets/catalog/oxford_flowers102}{oxford\_flowers102}
&  99.79  &  99.59
& train & val & test
& \hspace{-10pt}\citep{Nilsback2008AutomatedFC}
\\
\href{https://www.tensorflow.org/datasets/catalog/oxford_iiit_pet}{oxford\_iiit\_pet}
&  98.09  &  95.50
& train[{\tiny:2944}] & train[{\tiny2944:}] & test
& \citep{Parkhi2012CatsAD}
\\
\href{https://www.tensorflow.org/datasets/catalog/sun397}{sun397}
&  84.71  &  84.32
& train & val & test
& \citep{Xiao2010SUNDL}
\\
\href{https://www.tensorflow.org/datasets/catalog/svhn_cropped}{svhn\_cropped}
&  97.47  &  97.50
& train[{\tiny:65931}] & train[{\tiny65931:}] & test
& \citep{Netzer2011ReadingDI}
\\
\href{https://www.tensorflow.org/datasets/catalog/patch_camelyon}{patch\_camelyon}
&  92.82  &  91.14
& train & val & test
& \citep{Veeling2018RotationEC}
\\
\href{https://www.tensorflow.org/datasets/catalog/eurosat#eurosatrgb_default_config}{eurosat/rgb}
&  99.27  &  99.22
& train[{\tiny:16200}] & train[{\tiny16200:21600}] & train[{\tiny21600:}]
& \citep{Helber2019EuroSATAN}
\\
\href{https://www.tensorflow.org/datasets/catalog/resisc45}{resisc45}
&  97.87  &  97.02
& train[{\tiny:18900}] & train[{\tiny18900:25200}] & train[{\tiny25200:}]
& \citep{Cheng2017RemoteSI}
\\
\multicolumn{2}{l}{
% \href{https://www.tensorflow.org/datasets/catalog/diabetic_retinopathy_detection/#diabetic_retinopathy_detectionbtgraham-300}{diabetic\_retinopathy\_detection/...}
diabetic\_retinopathy\_detection/...
}
\\
\ \ \ \ \href{https://www.tensorflow.org/datasets/catalog/diabetic_retinopathy_detection/#diabetic_retinopathy_detectionbtgraham-300}{btgraham-300}
&  85.22  &  83.75
& train & val & test
& \hspace{-10pt}\citep{kaggle-diabetic-retinopathy}
\\
\href{https://www.tensorflow.org/datasets/catalog/clevr}{clevr/count\_cylinders}$^{[5]}$
&  99.65  &  99.47
& train[{\tiny:63000}] & train[{\tiny63000:}] & val
& \citep{Johnson2017CLEVRAD}
\\
\href{https://www.tensorflow.org/datasets/catalog/clevr}{clevr/count\_all}
&  99.97  &  99.90
& train[{\tiny:63000}] & train[{\tiny63000:}] & val
& \citep{Johnson2017CLEVRAD}
\\
\href{https://www.tensorflow.org/datasets/catalog/clevr}{clevr/closest\_object\_distance}
&  94.63  &  94.04
& train[{\tiny:63000}] & train[{\tiny63000:}] & val
& \citep{Johnson2017CLEVRAD}
\\
\href{https://www.tensorflow.org/datasets/catalog/dmlab}{dmlab}
&  77.01  &  76.91
& train & val & test
& \citep{Zhai2019TheVT}
\\
\href{https://www.tensorflow.org/datasets/catalog/dsprites}{dsprites/label\_x\_position}
&  99.99  &  99.98
& train[{\tiny:589824}] & train[{\tiny589824:663552}] & train[{\tiny663552:}]
& \citep{Klindt2021TowardsND}
\\
\href{https://www.tensorflow.org/datasets/catalog/dsprites}{dsprites/label\_orientation}
&  96.78  &  96.43
& train[{\tiny:589824}] & train[{\tiny589824:663552}] & train[{\tiny663552:}]
& \citep{Klindt2021TowardsND}
\\
\href{https://www.tensorflow.org/datasets/catalog/kitti}{kitti/closest\_object\_distance}$^{[5]}$
&  83.70  &  77.92
& train & val & test
& \citep{Geiger2012AreWR}
\\
\href{https://www.tensorflow.org/datasets/catalog/kitti}{kitti/count\_vehicles}$^{[5]}$
&  92.14  &  76.93
& train & val & test
& \citep{Geiger2012AreWR}
\\
\href{https://www.tensorflow.org/datasets/catalog/kitti}{kitti/closest\_vehicle\_distance}
&  89.76  &  82.28
& train & val & test
& \citep{Geiger2012AreWR}
\\
\href{https://www.tensorflow.org/datasets/catalog/smallnorb}{smallnorb/label\_category}$^{[5]}$
&  99.45  &  99.38
& train & test[{\tiny:50\%}] & test[{\tiny50\%:}]
& \citep{LeCun2004LearningMF}
\\
\href{https://www.tensorflow.org/datasets/catalog/smallnorb}{smallnorb/label\_lighting}$^{[5]}$
&  99.93  &  99.89
& train & test[{\tiny:50\%}] & test[{\tiny50\%:}]
& \citep{LeCun2004LearningMF}
\\
\href{https://www.tensorflow.org/datasets/catalog/smallnorb}{smallnorb/label\_azimuth}
&  35.40  &  34.44
& train & test[{\tiny:50\%}] & test[{\tiny50\%:}]
& \citep{LeCun2004LearningMF}
\\
\href{https://www.tensorflow.org/datasets/catalog/smallnorb}{smallnorb/label\_elevation}
&  96.22  &  96.26
& train & test[{\tiny:50\%}] & test[{\tiny50\%:}]
& \citep{LeCun2004LearningMF}
\\

\midrule

\multicolumn{7}{c}{Continues in Table~\ref{table:datasets2} \dots}  \\
\bottomrule
  \end{tabular}
\end{table*}

\begin{table*}[h]
\small
\caption{Datasets details (part 2 of 3).}
\label{table:datasets2}
\centering
\setlength\tabcolsep{1pt}
\hspace*{-9.82pt}
\begin{tabular}{lcccccc}
    \toprule
    & \multicolumn{2}{c}{Accuracy (\%)}
    & \multicolumn{3}{c}{Splits}                   \\
    \cmidrule(r){2-3}
    \cmidrule(r){4-6}
    Name & Val. & Test & Train & Val. & Test  & Reference\\
    \midrule
    % \midrule
\multicolumn{7}{c}{\dots Continues from Table~\ref{table:datasets} }\\
\midrule

\multicolumn{7}{c}{Visual domain decathlon benchmark} \\

\multicolumn{2}{l}{
% \href{https://www.tensorflow.org/datasets/catalog/visual_domain_decathlon#visual_domain_decathlonaircraft_default_config/}{visual\_domain\_decathlon/...}
visual\_domain\_decathlon/...
}
&
\\
\ \ \ \ \href{https://www.tensorflow.org/datasets/catalog/visual_domain_decathlon#visual_domain_decathlonimagenet12}{imagenet12}
&  89.47  &  89.70
& train & val[{\tiny:50\%}] & val[{\tiny50\%:}]
& \citep{hakanbilensylvestrerebuffitomasjakab2017}
\\
\ \ \ \ \href{https://www.tensorflow.org/datasets/catalog/visual_domain_decathlon#visual_domain_decathlonsvhn}{svhn}
&  98.75  &  98.57
& train & val[{\tiny:50\%}] & val[{\tiny50\%:}]
& \citep{hakanbilensylvestrerebuffitomasjakab2017}
\\
\ \ \ \ \href{https://www.tensorflow.org/datasets/catalog/visual_domain_decathlon#visual_domain_decathloncifar100}{cifar100}
&  97.79  &  97.98
& train & val[{\tiny:50\%}] & val[{\tiny50\%:}]
& \citep{hakanbilensylvestrerebuffitomasjakab2017}
\\
\ \ \ \ \href{https://www.tensorflow.org/datasets/catalog/visual_domain_decathlon#visual_domain_decathlongtsrb}{gtsrb}
&  99.97  &  99.95
& train & val[{\tiny:50\%}] & val[{\tiny50\%:}]
& \citep{hakanbilensylvestrerebuffitomasjakab2017}
\\
\ \ \ \ \href{https://www.tensorflow.org/datasets/catalog/visual_domain_decathlon#visual_domain_decathlondaimlerpedcls}{daimlerpedcls}
&  100  &  100
& train & val[{\tiny:50\%}] & val[{\tiny50\%:}]
& \citep{hakanbilensylvestrerebuffitomasjakab2017}
\\
\ \ \ \ \href{https://www.tensorflow.org/datasets/catalog/visual_domain_decathlon#visual_domain_decathlonomniglot}{omniglot}
&  88.70  &  88.05
& train & val[{\tiny:50\%}] & val[{\tiny50\%:}]
& \citep{hakanbilensylvestrerebuffitomasjakab2017}
\\
\ \ \ \ \href{https://www.tensorflow.org/datasets/catalog/visual_domain_decathlon#visual_domain_decathlonucf101}{ucf101}
&  85.96  &  87.70
& train & val[{\tiny:50\%}] & val[{\tiny50\%:}]
& \citep{hakanbilensylvestrerebuffitomasjakab2017}
\\
\ \ \ \ \href{https://www.tensorflow.org/datasets/catalog/visual_domain_decathlon#visual_domain_decathlonaircraft_default_config}{aircraft}
&  72.95  &  69.95
& train & val[{\tiny:50\%}] & val[{\tiny50\%:}]
& \citep{hakanbilensylvestrerebuffitomasjakab2017}
\\
\ \ \ \ \href{https://www.tensorflow.org/datasets/catalog/visual_domain_decathlon#visual_domain_decathlondtd}{dtd}
&  73.11  &  75.11
& train & val[{\tiny:50\%}] & val[{\tiny50\%:}]
& \citep{hakanbilensylvestrerebuffitomasjakab2017}
\\
\ \ \ \ \href{https://www.tensorflow.org/datasets/catalog/visual_domain_decathlon#visual_domain_decathlonvgg-flowers}{vgg-flowers}
&  99.58  &  99.41
& train & val[{\tiny:50\%}] & val[{\tiny50\%:}]
& \citep{hakanbilensylvestrerebuffitomasjakab2017}
\\
\midrule

\multicolumn{7}{c}{Multitask Character Classification Benchmark} \\

\href{https://www.tensorflow.org/datasets/catalog/emnist#emnistdigits}{emnist/digits}
&  99.86  &  99.81
& train[{\tiny5\%:}] & train[{\tiny:5\%}] & test
& \citep{Cohen2017EMNISTEM}
\\
\href{https://www.tensorflow.org/datasets/catalog/emnist#emnistletters}{emnist/letters}
&  96.57  &  95.03
&train[{\tiny5\%:}] & train[{\tiny:5\%}] & test
& \citep{Cohen2017EMNISTEM}
\\
\href{https://www.tensorflow.org/datasets/catalog/kmnist}{kmnist}
&  99.79  &  98.41
& train[{\tiny5\%:}] & train[{\tiny:5\%}] & test
& \citep{Clanuwat2018DeepLF}
\\
\href{https://www.tensorflow.org/datasets/catalog/mnist}{mnist}
&  99.82  &  99.71
& train[{\tiny5\%:}] & train[{\tiny:5\%}] & test
& \citep{LeCun1998GradientbasedLA}
\\
\href{https://www.tensorflow.org/datasets/catalog/omniglot}{omniglot}
&  100  &  100
& train & small1 & small2
& \citep{Lake2015HumanlevelCL}
\\
\href{https://www.tensorflow.org/datasets/catalog/cmaterdb#cmaterdbbangla_default_config}{cmaterdb/bangla}
&  99.89  &  99.00
& train[{\tiny20\%:}] & train[{\tiny:20\%}] & test
& \citep{Das2012AGA,Das2012ASF}
\\
\href{https://www.tensorflow.org/datasets/catalog/cmaterdb#cmaterdbdevanagari}{cmaterdb/devanagari}
&  100  &  97.00
& train[{\tiny20\%:}] & train[{\tiny:20\%}] & test
& \citep{Das2012AGA,Das2012ASF}
\\
\href{https://www.tensorflow.org/datasets/catalog/cmaterdb#cmaterdbtelugu}{cmaterdb/telugu}
&  100  &  99.40
& train[{\tiny20\%:}] & train[{\tiny:20\%}] & test
& \citep{Das2012AGA,Das2012ASF}
\\

\midrule

\multicolumn{7}{c}{VTAB-1k benchmark$^{[4]}$} \\

\href{https://www.tensorflow.org/datasets/catalog/caltech101}{caltech101}
&  98.97  &  89.91
& train[{\tiny:800}] & train[{\tiny2754:2954}] & test
& \citep{FeiFei2004LearningGV}
\\
\href{https://www.tensorflow.org/datasets/catalog/cifar100}{cifar100}
&  96.92  &  92.85
& train[{\tiny:800}] & train[{\tiny45000:45200}] & test
& \citep{Krizhevsky2009LearningML}
\\
\href{https://www.tensorflow.org/datasets/catalog/cifar10}{cifar10}
&  100  &  99.31
& train[{\tiny:800}] & train[{\tiny45000:45200}] & test
& \citep{Krizhevsky2009LearningML}
\\
\href{https://www.tensorflow.org/datasets/catalog/dtd}{dtd}
&  82.05  &  77.87
& train[{\tiny:800}] & val[{\tiny:200}] & test
& \citep{Cimpoi2014DescribingTI}
\\
\href{https://www.tensorflow.org/datasets/catalog/oxford_flowers102}{oxford\_flowers102}
&  100  &  99.32
& train[{\tiny:800}] & val[{\tiny:200}] & test
& \hspace{-10pt}\citep{Nilsback2008AutomatedFC}
\\
\href{https://www.tensorflow.org/datasets/catalog/oxford_iiit_pet}{oxford\_iiit\_pet}
&  97.44  &  93.51
& train[{\tiny:800}] & train[{\tiny2944:3144}] & test
& \citep{Parkhi2012CatsAD}
\\
\href{https://www.tensorflow.org/datasets/catalog/sun397}{sun397}
&  66.67  &  60.94
& train[{\tiny:800}] & val[{\tiny:200}] & test
& \citep{Xiao2010SUNDL}
\\
\href{https://www.tensorflow.org/datasets/catalog/svhn_cropped}{svhn\_cropped}
&  98.00  &  97.46
& train[{\tiny:800}] & train[{\tiny65931:66131}] & test
& \citep{Netzer2011ReadingDI}
\\
\href{https://www.tensorflow.org/datasets/catalog/patch_camelyon}{patch\_camelyon}
&  96.41  &  91.57
& train[{\tiny:800}] & val[{\tiny:200}] & test
& \citep{Veeling2018RotationEC}
\\
\href{https://www.tensorflow.org/datasets/catalog/eurosat#eurosatrgb_default_config}{eurosat/rgb}
 &  99.49  &  98.56
& train[{\tiny:800}] & train[{\tiny16200:16400}] & train[{\tiny21600:}]
& \citep{Helber2019EuroSATAN}
\\
\href{https://www.tensorflow.org/datasets/catalog/resisc45}{resisc45}
&  97.50  &  95.37
& train[{\tiny:800}] & train[{\tiny18900:19100}] & train[{\tiny25200:}]
& \citep{Cheng2017RemoteSI}
\\
\multicolumn{2}{l}{
% \href{https://www.tensorflow.org/datasets/catalog/diabetic_retinopathy_detection/#diabetic_retinopathy_detectionbtgraham-300}{
diabetic\_retinopathy\_detection/...
% }
}
\\
\ \ \ \ \href{https://www.tensorflow.org/datasets/catalog/diabetic_retinopathy_detection/#diabetic_retinopathy_detectionbtgraham-300}{btgraham-300}
&  88.50  &  82.77
& train[{\tiny:800}] & val[{\tiny:200}] & test
& \hspace{-10pt}\citep{kaggle-diabetic-retinopathy}
\\
\href{https://www.tensorflow.org/datasets/catalog/clevr}{clevr/count\_cylinders}$^{[5]}$
 &  99.49  &  99.01
& train[{\tiny:800}] & train[{\tiny63000:63200}] & val
& \citep{Johnson2017CLEVRAD}
\\
\href{https://www.tensorflow.org/datasets/catalog/clevr}{clevr/count\_all}
&  100  &  99.88
& train[{\tiny:800}] & train[{\tiny63000:63200}] & val
& \citep{Johnson2017CLEVRAD}
\\
\href{https://www.tensorflow.org/datasets/catalog/clevr}{clevr/closest\_object\_distance}
&  92.31  &  92.39
& train[{\tiny:800}] & train[{\tiny63000:63200}] & val
& \citep{Johnson2017CLEVRAD}
\\
\href{https://www.tensorflow.org/datasets/catalog/dmlab}{dmlab}
&  77.44  &  74.70
& train[{\tiny:800}] & val[{\tiny:200}] & test
& \citep{Zhai2019TheVT}
\\
\href{https://www.tensorflow.org/datasets/catalog/dsprites}{dsprites/label\_x\_position}
&  100  &  99.44
& train[{\tiny:800}] & train[{\tiny589824:590024}] & train[{\tiny663552:}]
& \citep{Klindt2021TowardsND}
\\
\href{https://www.tensorflow.org/datasets/catalog/dsprites}{dsprites/label\_orientation}
&  97.50  &  96.29
& train[{\tiny:800}] & train[{\tiny589824:590024}] & train[{\tiny663552:}]
& \citep{Klindt2021TowardsND}
\\
\href{https://www.tensorflow.org/datasets/catalog/kitti}{kitti/closest\_object\_distance}$^{[5]}$
&  84.50  &  78.34
& train[{\tiny:800}] & val[{\tiny:200}] & test
& \citep{Geiger2012AreWR}
\\
\href{https://www.tensorflow.org/datasets/catalog/kitti}{kitti/count\_vehicles}$^{[5]}$
&  93.85  &  69.34
& train[{\tiny:800}] & val[{\tiny:200}] & test
& \citep{Geiger2012AreWR}
\\
\href{https://www.tensorflow.org/datasets/catalog/kitti}{kitti/closest\_vehicle\_distance}
&  88.00  &  82.14
& train[{\tiny:800}] & val[{\tiny:200}] & test
& \citep{Geiger2012AreWR}
\\
\href{https://www.tensorflow.org/datasets/catalog/smallnorb}{smallnorb/label\_category}$^{[5]}$
&  100  &  97.66
& train[{\tiny:800}] & test[{\tiny:200}] & test[{\tiny50\%:}]
& \citep{LeCun2004LearningMF}
\\
\href{https://www.tensorflow.org/datasets/catalog/smallnorb}{smallnorb/label\_lighting}$^{[5]}$
 &  100  &  98.29
& train[{\tiny:800}] & test[{\tiny:200}] & test[{\tiny50\%:}]
& \citep{LeCun2004LearningMF}
\\
\href{https://www.tensorflow.org/datasets/catalog/smallnorb}{smallnorb/label\_azimuth}
&  37.00  &  33.72
& train[{\tiny:800}] & test[{\tiny:200}] & test[{\tiny50\%:}]
& \citep{LeCun2004LearningMF}
\\
\href{https://www.tensorflow.org/datasets/catalog/smallnorb}{smallnorb/label\_elevation}
&  92.50  &  92.81
& train[{\tiny:800}] & test[{\tiny:200}] & test[{\tiny50\%:}]
& \citep{LeCun2004LearningMF}
\\
\midrule
\multicolumn{7}{c}{Continues in Table~\ref{table:datasets3} \dots}  \\

    \bottomrule
  \end{tabular}
\end{table*}

\clearpage

\begin{table*}[h]
\small
\caption{Datasets details (part 3 of 3).}
\label{table:datasets3}
\centering
\setlength\tabcolsep{1pt}
\begin{tabular}{lcccccc}
    \toprule
    & \multicolumn{2}{c}{Accuracy (\%)}
    & \multicolumn{3}{c}{Splits}                   \\
    \cmidrule(r){2-3}
    \cmidrule(r){4-6}
    Name & Val. & Test & Train & Val. & Test  & Reference\\
    \midrule
\multicolumn{7}{c}{\dots Continues from Table~\ref{table:datasets2} }\\
\midrule
\midrule
\multicolumn{7}{c}{\textbf{Task-set B}}\\
\midrule

\href{https://www.tensorflow.org/datasets/catalog/beans}{beans}
&  100  &  96.88
 & train & val & test & 
 \citep{beansdata}
 \\
\href{https://www.tensorflow.org/datasets/catalog/binary_alpha_digits}{binary\_alpha\_digits}
&  91.43  &  85.71
 & train[{\tiny10\%:}] & train[{\tiny5\%:10\%}] & train[{\tiny:5\%}] &
 $-$ %  Missing \citep{ }
 \\
\href{https://www.tensorflow.org/datasets/catalog/caltech_birds2010}{caltech\_birds2010}
&  98.00  &  89.52
 & train[{\tiny5\%:}] & train[{\tiny:5\%}] & test & 
 \citep{Welinder2010CaltechUCSDB2}
 \\
\href{https://www.tensorflow.org/datasets/catalog/caltech_birds2011}{caltech\_birds2011}
&  93.00  &  90.99
 & train[{\tiny5\%:}] & train[{\tiny:5\%}] & test & 
 \citep{Welinder2010CaltechUCSDB2}
 \\
\href{https://www.tensorflow.org/datasets/catalog/cars196}{cars196}
&  91.79  &  90.41
 & train[{\tiny5\%:}] & train[{\tiny:5\%}] & test & 
 \citep{Krause20133DOR}
 \\
\href{https://www.tensorflow.org/datasets/catalog/cassava}{cassava}
&  92.31  &  91.72
 & train & val & test & 
 \citep{Mwebaze2019iCassava2V}
 \\
\href{https://www.tensorflow.org/datasets/catalog/cats_vs_dogs}{cats\_vs\_dogs}
&  100  &  99.91
 & train[{\tiny10\%:}] & train[{\tiny5\%:10\%}] & train[{\tiny:5\%}] &
 \citep{Elson2007AsirraAC}
 \\
\href{https://www.tensorflow.org/datasets/catalog/citrus_leaves}{citrus\_leaves}
&  100  &  93.33
 & train[{\tiny10\%:}] & train[{\tiny5\%:10\%}] & train[{\tiny:5\%}] &
 \citep{Rauf2019ACF}
 \\
\href{https://www.tensorflow.org/datasets/catalog/colorectal_histology}{colorectal\_histology}
&  99.17  &  98.40
 & train[{\tiny10\%:}] & train[{\tiny5\%:10\%}] & train[{\tiny:5\%}] &
 \citep{Kather2016MulticlassTA}
 \\
\multicolumn{2}{l}{
% \href{https://www.tensorflow.org/datasets/catalog/controlled_noisy_web_labels}{
controlled\_noisy\_web\_labels/...
% }
}
 &
 \\
\ \ \ \ \href{https://www.tensorflow.org/datasets/catalog/controlled_noisy_web_labels#mini_imagenet_red}{mini\_imagenet\_red}
&  96.35  &  95.12
 & train\_00 & val[{\tiny:50\%}] & val[{\tiny50\%:}] & 
 \citep{Jiang2020BeyondSN}
 \\
\ \ \ \ \href{https://www.tensorflow.org/datasets/catalog/controlled_noisy_web_labels#mini_imagenet_blue}{mini\_imagenet\_blue}
&  96.52  &  95.60
 & train\_00 & val[{\tiny:50\%}] & val[{\tiny50\%:}] &  \citep{Jiang2020BeyondSN}
 \\
\multicolumn{2}{l}{
% \href{https://www.tensorflow.org/datasets/catalog/curated_breast_imaging_ddsm#patches}{
curated\_breast\_imaging\_ddsm/...
% }
}
 &
 \\
\ \ \ \ \href{https://www.tensorflow.org/datasets/catalog/curated_breast_imaging_ddsm#patches}{patches}
&  70.66  &  67.51
 & train & val & test & 
 \citep{Clark2013TheCI}
 \\
% \href{https://www.tensorflow.org/datasets/catalog/cycle_gan#apple2orange}{
cycle\_gan/...
% }
 &
 \\
\ \ \ \ \href{https://www.tensorflow.org/datasets/catalog/cycle_gan#apple2orange}{apple2orange}
 &  100  &  98.83
 &  {\tiny trainA+B[10\%:]} & {\tiny trainA+B[:10\%]} & {\tiny testA+B}&  
 \citep{Zhu2017UnpairedIT}
 \\
\ \ \ \ \href{https://www.tensorflow.org/datasets/catalog/cycle_gan#summer2winter_yosemite}{summer2winter} %\_yosemite}
 &  95.71  &  91.22
 &  {\tiny trainA+B[10\%:]} & {\tiny trainA+B[:10\%]} & {\tiny testA+B}&
 \citep{Zhu2017UnpairedIT}
 \\
\ \ \ \ \href{https://www.tensorflow.org/datasets/catalog/cycle_gan#horse2zebra}{horse2zebra}
&  100  &  99.23
 &  {\tiny trainA+B[10\%:]} & {\tiny trainA+B[:10\%]} & {\tiny testA+B}&
 \citep{Zhu2017UnpairedIT}
 \\
\ \ \ \ \href{https://www.tensorflow.org/datasets/catalog/cycle_gan#monet2photo}{monet2photo}
&  100  &  100
 &  {\tiny trainA+B[10\%:]} & {\tiny trainA+B[:10\%]} & {\tiny testA+B}&
 \citep{Zhu2017UnpairedIT}
 \\
\ \ \ \ \href{https://www.tensorflow.org/datasets/catalog/cycle_gan#cezanne2photo}{cezanne2photo}
&  100  &  100
 &  {\tiny trainA+B[10\%:]} & {\tiny trainA+B[:10\%]} & {\tiny testA+B}&
 \citep{Zhu2017UnpairedIT}
 \\
\ \ \ \ \href{https://www.tensorflow.org/datasets/catalog/cycle_gan#ukiyoe2photo}{ukiyoe2photo}
&  100  &  99.70
 &  {\tiny trainA+B[10\%:]} & {\tiny trainA+B[:10\%]} & {\tiny testA+B}&
 \citep{Zhu2017UnpairedIT}
 \\
\ \ \ \ \href{https://www.tensorflow.org/datasets/catalog/cycle_gan#vangogh2photo}{vangogh2photo}
&  100  &  100
 &  {\tiny trainA+B[10\%:]} & {\tiny trainA+B[:10\%]} & {\tiny testA+B}&
 \citep{Zhu2017UnpairedIT}
 \\
\ \ \ \ \href{https://www.tensorflow.org/datasets/catalog/cycle_gan#maps}{maps}
&  100  &  100
 &  {\tiny trainA+B[10\%:]} & {\tiny trainA+B[:10\%]} & {\tiny testA+B}&
 \citep{Zhu2017UnpairedIT}
 \\
\ \ \ \ \href{https://www.tensorflow.org/datasets/catalog/cycle_gan#cityscapes}{cityscapes}
&  100  &  100
 &  {\tiny trainA+B[10\%:]} & {\tiny trainA+B[:10\%]} & {\tiny testA+B}&
 \citep{Zhu2017UnpairedIT}
 \\
\ \ \ \ \href{https://www.tensorflow.org/datasets/catalog/cycle_gan#facades}{facades}
&  100  &  100
 &  {\tiny trainA+B[10\%:]} & {\tiny trainA+B[:10\%]} & {\tiny testA+B}&
 \citep{Zhu2017UnpairedIT}
 \\
\ \ \ \ \href{https://www.tensorflow.org/datasets/catalog/cycle_gan#iphone2dslr_flower}{iphone2dslr\_flower}
&  98.99  &  94.66
 &  {\tiny trainA+B[10\%:]} & {\tiny trainA+B[:10\%]} & {\tiny testA+B}&
 \citep{Zhu2017UnpairedIT}
 \\
\href{https://www.tensorflow.org/datasets/catalog/deep_weeds}{deep\_weeds}
&  98.79  &  98.06
 & train[{\tiny10\%:}] & train[{\tiny5\%:10\%}] & train[{\tiny:5\%}] &
 \citep{Olsen2019DeepWeedsAM}
 \\
\href{https://www.tensorflow.org/datasets/catalog/domainnet#real}{domainnet/real}
&  90.90  &  90.23
 & train[{\tiny5\%:}] & train[{\tiny:5\%}] & test &
 \citep{Peng2019MomentMF}
 \\
\href{https://www.tensorflow.org/datasets/catalog/domainnet#painting}{domainnet/painting}
&  82.62  &  82.11
 & train[{\tiny5\%:}] & train[{\tiny:5\%}] & test &
 \citep{Peng2019MomentMF}
 \\
\href{https://www.tensorflow.org/datasets/catalog/domainnet#clipart}{domainnet/clipart}
&  87.16  &  85.35
 & train[{\tiny5\%:}] & train[{\tiny:5\%}] & test &
 \citep{Peng2019MomentMF}
 \\
\href{https://www.tensorflow.org/datasets/catalog/domainnet#quickdraw}{domainnet/quickdraw}
&  76.46  &  76.50
 & train[{\tiny5\%:}] & train[{\tiny:5\%}] & test & 
 \citep{Peng2019MomentMF}
 \\
\href{https://www.tensorflow.org/datasets/catalog/domainnet#infograph}{domainnet/infograph}
&  56.28  &  55.31
& train[{\tiny5\%:}] & train[{\tiny:5\%}] & test &
 \citep{Peng2019MomentMF}
 \\
\href{https://www.tensorflow.org/datasets/catalog/domainnet#sketch}{domainnet/sketch}
&  79.96  &  79.51
 & train[{\tiny5\%:}] & train[{\tiny:5\%}] & test &
 \citep{Peng2019MomentMF}
 \\
\href{https://www.tensorflow.org/datasets/catalog/food101}{food101}
&  94.58  &  91.47
 & train[{\tiny5\%:}] & val & train[{\tiny:5\%}] &
 \citep{Bossard2014Food101M}
 \\
\href{https://www.tensorflow.org/datasets/catalog/horses_or_humans}{horses\_or\_humans}
&  100  &  99.61
 & train[{\tiny5\%:}] & train[{\tiny:5\%}] & test &
 \citep{horses_or_humans}
 \\
\href{https://www.tensorflow.org/datasets/catalog/i_naturalist2017}{i\_naturalist2017}
&  77.83  &  78.19
 & train & val[{\tiny:50\%}] & val[{\tiny50\%:}] & 
 \citep{Horn2018TheIS}
 \\
\href{https://www.tensorflow.org/datasets/catalog/i_naturalist2018}{i\_naturalist2018}
&  81.37  &  81.63
 & train & val[{\tiny:50\%}] & val[{\tiny50\%:}] & 
 \citep{Horn2018TheIS}
 \\
\href{https://www.tensorflow.org/datasets/catalog/imagenet_a}{imagenet\_a}
&  90.28  &  88.00
 & train[{\tiny10\%:}] & train[{\tiny5\%:10\%}] & train[{\tiny:5\%}] & 
 \citep{Hendrycks2021NaturalAE}
 \\
\href{https://www.tensorflow.org/datasets/catalog/imagenet_lt}{imagenet\_lt}
&  87.59  &  82.68
 & train & val & test & 
 \citep{Liu2019LargeScaleLR}
 \\
\href{https://www.tensorflow.org/datasets/catalog/imagenet_r}{imagenet\_r}
&  91.13  &  89.87
 & train[{\tiny10\%:}] & train[{\tiny5\%:10\%}] & train[{\tiny:5\%}] & 
 \citep{Hendrycks2021TheMF}
 \\
\href{https://www.tensorflow.org/datasets/catalog/imagenet_sketch}{imagenet\_sketch}
&  89.67  &  88.52
 & train[{\tiny10\%:}] & train[{\tiny5\%:10\%}] & train[{\tiny:5\%}] & 
 \citep{Wang2019LearningRG}
 \\
\href{https://www.tensorflow.org/datasets/catalog/imagenette}{imagenette}
&  99.92  &  100
 & train[{\tiny5\%:}] & val & train[{\tiny:5\%}] & 
 \citep{imagenette}
 \\
\href{https://www.tensorflow.org/datasets/catalog/imagewang}{imagewang}
&  97.32  &  99.59
 & train[{\tiny5\%:}] & val & train[{\tiny:5\%}] & 
 \citep{imagewang}
 \\
\href{https://www.tensorflow.org/datasets/catalog/malaria}{malaria}
&  98.39  &  97.68
 & train[{\tiny10\%:}] & train[{\tiny5\%:10\%}] & train[{\tiny:5\%}] &
 \citep{Rajaraman2018PretrainedCN}
 \\
\href{https://www.tensorflow.org/datasets/catalog/pet_finder}{pet\_finder}
&  62.40  &  60.77
 & train[{\tiny10\%:}] & train[{\tiny5\%:10\%}] & train[{\tiny:5\%}] &
 $-$ %\citep{ }
 \\
\href{https://www.tensorflow.org/datasets/catalog/places365_small}{places365\_small}
&  59.37  &  59.85
 & train & val[{\tiny:50\%}] & val[{\tiny50\%:}] &
 \citep{Zhou2018PlacesA1}
 \\
\href{https://www.tensorflow.org/datasets/catalog/plant_village}{plant\_village}
&  100  &  99.89
 & train[{\tiny10\%:}] & train[{\tiny5\%:10\%}] & train[{\tiny:5\%}] &
 \citep{Hughes2015AnOA}
 \\
\href{https://www.tensorflow.org/datasets/catalog/plantae_k}{plantae\_k}
&  99.05  &  90.74
 & train[{\tiny10\%:}] & train[{\tiny5\%:10\%}] & train[{\tiny:5\%}] &
 \citep{Kour2019PlantaeKAL}
 \\
\href{https://www.tensorflow.org/datasets/catalog/quickdraw_bitmap}{quickdraw\_bitmap}
&  80.79  &  79.99
 & train[{\tiny20k:}] & train[{\tiny10k:20k}] & train[{\tiny:10k}] & 
 \citep{Ha2018ANR}
 \\
\href{https://www.tensorflow.org/datasets/catalog/rock_paper_scissors}{rock\_paper\_scissors}
&  100  &  97.31
 & train[{\tiny5\%:}] & train[{\tiny:5\%}] & test &
 \citep{rps}
 \\
\href{https://www.tensorflow.org/datasets/catalog/siscore#rotation}{siscore/rotation}
&  100  &  100
 & train[{\tiny10\%:}] & train[{\tiny5\%:10\%}] & train[{\tiny:5\%}] &
 \citep{Djolonga2021OnRA}
 \\
\href{https://www.tensorflow.org/datasets/catalog/siscore#size}{siscore/size}
&  99.93  &  99.94
 & train[{\tiny10\%:}] & train[{\tiny5\%:10\%}] & train[{\tiny:5\%}] &
 \citep{Djolonga2021OnRA}
 \\
\href{https://www.tensorflow.org/datasets/catalog/siscore#location}{siscore/location}
&  99.99  &  99.97
 & train[{\tiny10\%:}] & train[{\tiny5\%:10\%}] & train[{\tiny:5\%}] &
 \citep{Djolonga2021OnRA}
 \\
\href{https://www.tensorflow.org/datasets/catalog/stanford_dogs}{stanford\_dogs}
&  95.17  &  93.50
 & train[{\tiny5\%:}] & train[{\tiny:5\%}] & test &
 \citep{Khosla2012NovelDF}
 \\
\href{https://www.tensorflow.org/datasets/catalog/stanford_online_products}{stanford\_online\_products} %\hspace{-20pt}
&  90.00  &  89.49
 & train & test[{\tiny:10k}] & test[{\tiny10k:}] & 
 \citep{Song2016DeepML}
 \\
\href{https://www.tensorflow.org/datasets/catalog/stl10}{stl10}
&  100  &  99.60
 & train[{\tiny5\%:}] & train[{\tiny:5\%}] & test &
 \citep{Coates2011AnAO}
 \\
\href{https://www.tensorflow.org/datasets/catalog/tf_flowers}{tf\_flowers}
&  99.45  &  97.83
 & train[{\tiny10\%:}] & train[{\tiny5\%:10\%}] & train[{\tiny:5\%}] &
 $-$ %\citep{ }
 \\
\href{https://www.tensorflow.org/datasets/catalog/uc_merced}{uc\_merced}
&  100  &  100
 & train[{\tiny10\%:}] & train[{\tiny5\%:10\%}] & train[{\tiny:5\%}] &
 \citep{Yang2010BagofvisualwordsAS}
 \\

    \bottomrule
  \end{tabular}
\end{table*}

\clearpage

\begin{figure}[h]
\centering
\includegraphics[width=0.65\linewidth]{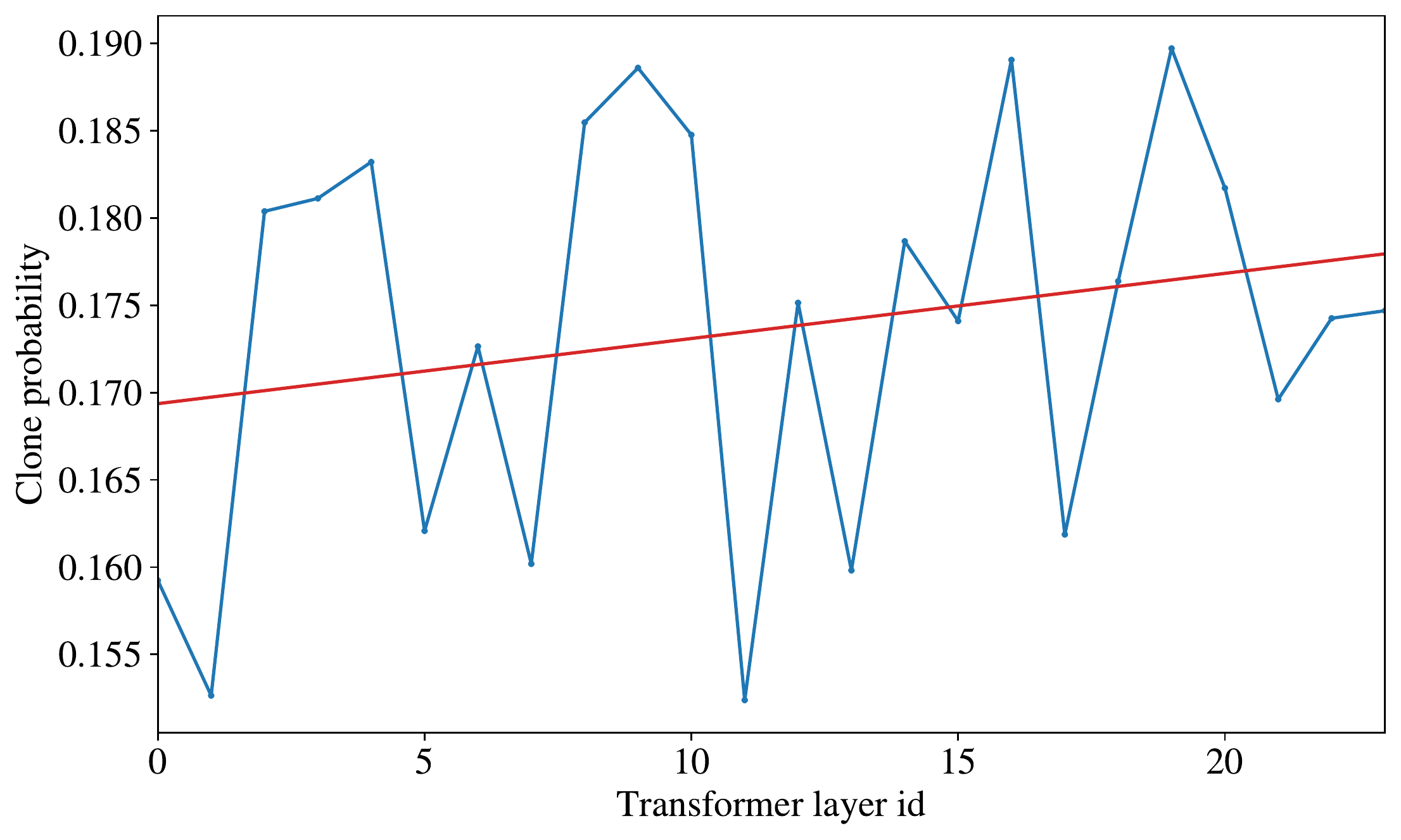}
\caption{Displays the mutation probabilities learned by the $\mu$ function for cloning transformer layers at different depth, lower layer ids correspond to transformer layers closer to the model input.
The displayed mutation probabilities are averaged across the values set by the $\mu$ functions learned for the 124 models included in the multitask system at the end of the sequence of experiment described in Section~\ref{section:experiments2}.
The red line is fitted to minimize the squared distance to the curve.
}
\label{fig:clone-prob}
\end{figure}

\begin{figure}[h]
\centering
\includegraphics[width=1.\linewidth]{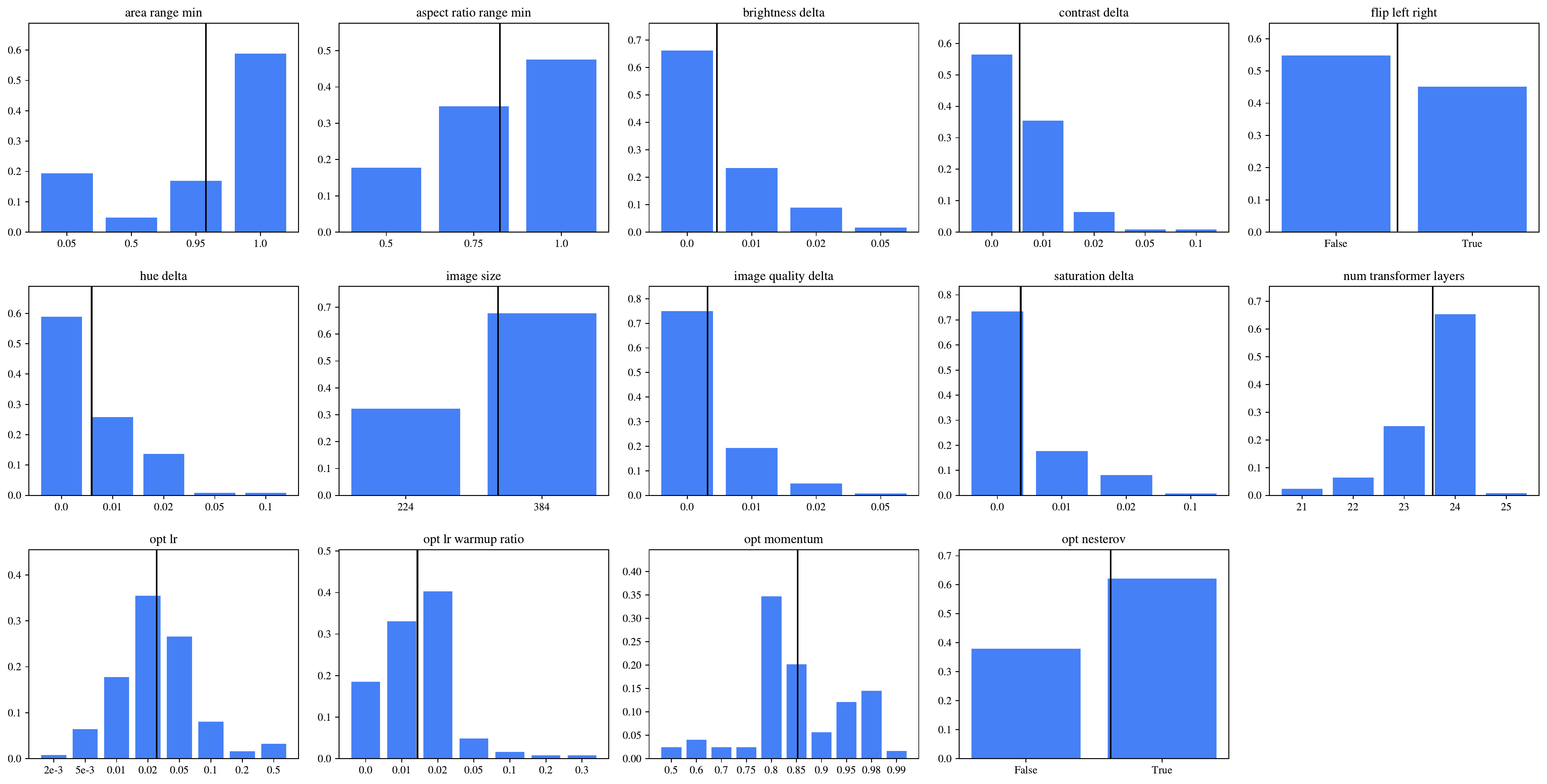}
\caption{Distributions of the hyperparameter values used by the 124 models included in the multitask system at the end of the sequence of experiment described in Section~\ref{section:experiments2}. The vertical black lines displays the average value.
}
\label{fig:dists}
\end{figure}

\begin{figure}[h]
\centering
\includegraphics[width=1.\linewidth]{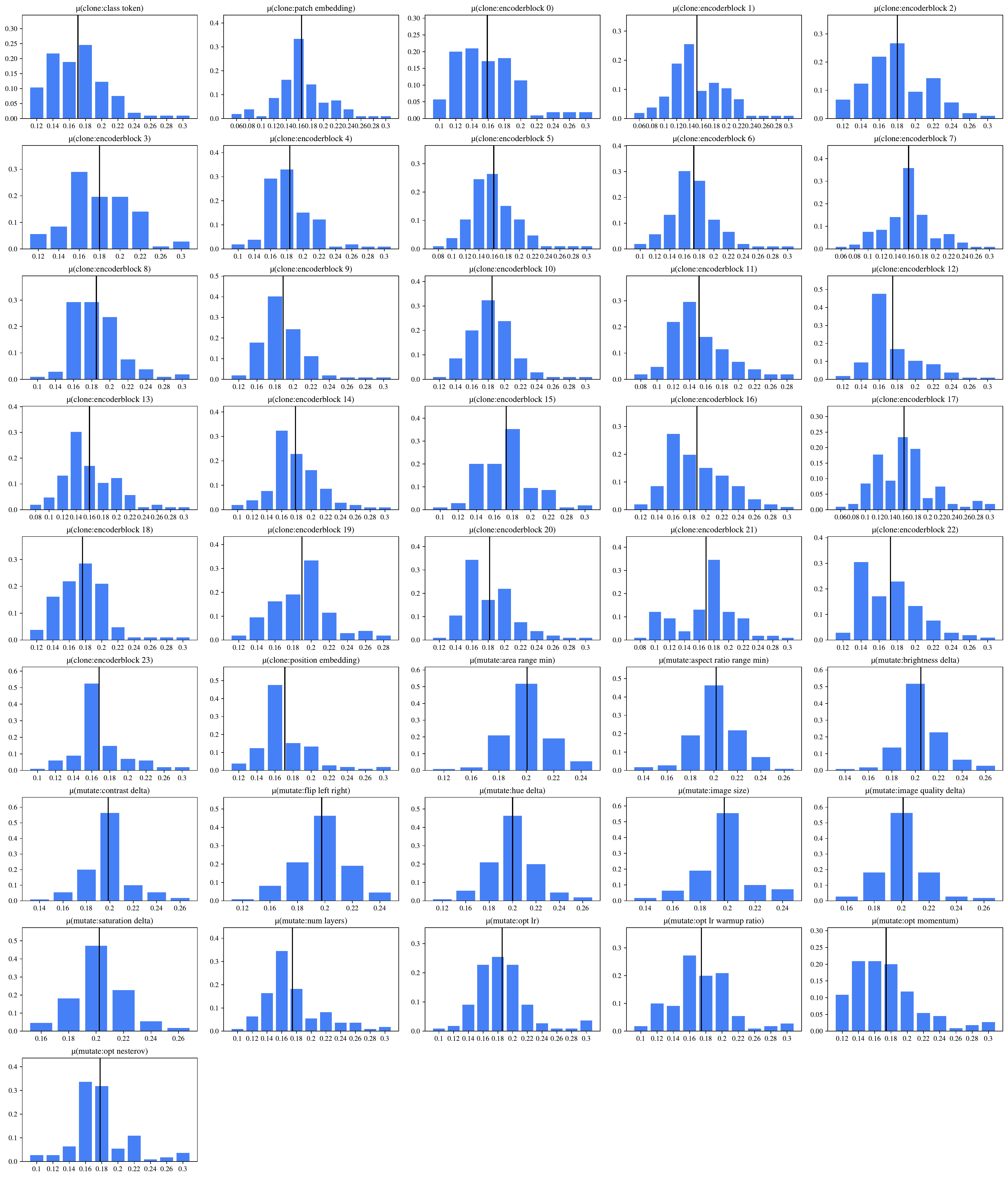}
\caption{Distributions of the $\mu(\cdot)$ values conditioned on different mutations of hyperparameters and clonable layers.
The histograms aggregates the values of the 124 models included in the multitask system at the end of the sequence of experiment described in Section~\ref{section:experiments2}. 
}
\label{fig:dists_mu}
\end{figure}

% \begin{ack}
% Ack.
% \end{ack}

\clearpage
% \medskip
% \section*{References}
% {
% \small
% https://www.overleaf.com/learn/latex/Natbib_bibliography_styles
% \bibliographystyle{abbrvnat} %plain} %plainnat} %abbrvnat}
% \bibliography{neurips_2022}
\bibliography{iclr2023_conference}
\bibliographystyle{iclr2023_conference}
% }

%%%%%%%%%%%%%%%%%%%%%%%%%%%%%%%%%%%%%%%%%%%%%%%%%%%%%%%%%%%%

\clearpage
\appendix

% \section{Appendix}

\end{document}